\documentclass[conference]{IEEEtran}
\IEEEoverridecommandlockouts
\usepackage{cite}
\usepackage{amsmath,amssymb,amsfonts}
\usepackage{algorithmic}
\usepackage{graphicx}
\usepackage{textcomp}
\usepackage{xcolor}
\usepackage{subfig}
\usepackage{widetext}
\usepackage{rotating}
\usepackage{float}

\def\BibTeX{{\rm B\kern-.05em{\sc i\kern-.025em b}\kern-.08em
    T\kern-.1667em\lower.7ex\hbox{E}\kern-.125emX}}
\begin{document}

\title{Occlusion-Robust Online Multi-Object Visual Tracking using a GM-PHD Filter with CNN-Based Re-Identification\\
}

\author{\IEEEauthorblockN{Nathanael L. Baisa*,~\IEEEmembership{Member,~IEEE,}}
\thanks{*Nathanael L. Baisa is with the School of Computer Science and Informatics, De Montfort University, Leicester LE1 9BH, UK. Some part of this work was done when the author was with the School of Computer Science, University of Lincoln, Lincoln LN6 7TS, United Kingdom. (e-mail: nathanaellmss@gmail.com).}}

\maketitle

\begin{abstract}
We propose a novel online multi-object visual tracker using a Gaussian mixture Probability Hypothesis Density (GM-PHD) filter and deep appearance learning. The GM-PHD filter has a linear complexity with the number of objects and observations while estimating the states and cardinality of time-varying number of objects, however, it is susceptible to miss-detections and does not include the identity of objects. We use visual-spatio-temporal information obtained from object bounding boxes and deeply learned appearance representations to perform estimates-to-tracks data association for target labeling as well as formulate an augmented likelihood and then integrate into the update step of the GM-PHD filter. We also employ additional unassigned tracks prediction after the data association step to overcome the susceptibility of the GM-PHD filter towards miss-detections caused by occlusion. Extensive evaluations on MOT16, MOT17 and HiEve benchmark datasets show that our tracker significantly outperforms several state-of-the-art trackers in terms of tracking accuracy and identification.

\end{abstract}

\begin{IEEEkeywords}
Online visual tracking, GM-PHD filter, Prediction, CNN features, Augmented likelihood, Re-identification
\end{IEEEkeywords}

\section{Introduction}

Multi-target tracking is an active research field in computer vision with a wide variety of applications such as intelligent surveillance, autonomous driving, robot navigation and augmented reality. Its main purpose is to estimate the states (e.g. locations) of objects from noisy detections, recognize their identities in each video frame and produce their trajectories. The most commonly adopted paradigm for multi-target tracking in computer vision is tracking-by-detection. This is due to the remarkable advances made in object detection algorithms driven by deep learning. In this tracking-by-detection paradigm, multi-target filters and/or data association are applied to object detections obtained from the object detector(s) applied to video frames to generate trajectories of tracked targets over-time. To perform this, online~\cite{MatPoiCav16}\cite{SonJeo16} and offline (batch)~\cite{LeaCanSch16}\cite{MilRotSch14}\cite{PirRamFow11} tracking approaches are the commonly used ones in the literature. The online tracking methods estimate the target state using Bayesian filtering at each time instant using current detections and rely on prediction to handle miss-detections using motion models to continue tracking. However, both past and future detections are fed into mainly global optimization-based data association approaches to handle miss-detections in offline tracking methods. Generally, the offline tracking approaches outperform the online tracking methods though they are limited for time-critical real-time applications where it is crucial to provide state estimates as the detections arrive such as in autonomous driving and robot navigation.

Multi-target tracking algorithms generally receive a random number of detections when object detector is applied to a video frame. When the object detector is applied to this video frame, there can be information uncertainty usually considered as measurement origin uncertainty~\cite{VoMalBarCorOsbMahVo15} which include miss-detection, clutter and very near unresolved objects. Thus, in addition to this measurement origin uncertainty, the multi-object tracking method needs to handle the targets' births, deaths, and the process and observation noises. As surveyed in~\cite{LuoZhaKim14}\cite{VoMalBarCorOsbMahVo15}, the three commonly known traditional data association methods used for numerous applications are Global Nearest Neighbor (GNN)~\cite{BarWilTia11}, Joint Probabilistic Data Association Filter (JPDAF)~\cite{BarWilTia11} and Multiple Hypothesis Tracking (MHT)~\cite{KimLiCip15,BarWilTia11}. While the GNN (computed using the Hungarian algorithm~\cite{FraJea71}) is sensitive to noise, the JPDAF and the MHT are computationally very expensive. Since these methods are computationally expensive and heavily rely on heuristics to track time-varying number of objects, another multi-target tracking approach has been proposed based on random finite set (RFS) theory~\cite{Mah14}. This approach includes all sources of uncertainty in a unified probabilistic framework. A probability hypothesis density (PHD) filter~\cite{Mah03} is the most commonly adopted RFS-based filter in computer vision for tracking targets in video sequences since it has a linear complexity with the number of objects and observations.

The PHD filter allows target birth, death, clutter (false alarms), and missing detections, however, it does not naturally incorporate the identity of objects in the framework since it is based on indistinguishability assumption of the point process. In order to include the identity of objects, additional technique is needed. This filter is also very susceptible to miss-detections. In fact, the PHD filter is designed originally for radar tracking applications where observations collected can contain numerous false alarms with very few miss-detections. However, in visual tracking applications, observations obtained from the recent deep learning-driven object detectors can contain very low false alarms (false positives) with high level of miss-detections (false negatives) due to occlusion. The parameter which controls the detection and miss-detection part of the PHD filter is the probability of detection ($p_D$, see in Section~\ref{Sec:GMPHD-Filter} in the Gaussian mixture implementation of the PHD (GM-PHD) filter). In my experiment, The GM-PHD filter~\cite{VoMa06} works if $p_D$ is set to not less than about 0.8 unless the covariance matrix fails to be a square, symmetric, positive definite matrix which in turn forces the GM-PHD filter to crash. This means even if we set $p_D$ to 0.8, the miss-detected target can not be maintained since the probability of detection drops too quickly (probability of miss-detection $p_{MD}= 1.0 - 0.8 = 0.2$). This is referred to as target death problem where targets die faster than they should when a miss-detection happens. Thus, naturally the GM-PHD filter is robust to false positives but it is very susceptible to miss-detections.

More recently, outstanding results have been obtained on a wide range of tasks using deep Convolutional Neural Network (CNN) features such as object recognition~\cite{KriHin12}\cite{KaiXiaSha15}, object detection~\cite{ShaKaiRos15} and person re-identification~\cite{KaiTao19}. Better performance has also been obtained on multi-target tracking using deep learning~\cite{LeaCanSch16,AmiAleSil17} since deeply learned appearance representations of objects have a capability of discriminating object of interest from not only background but also other objects of similar appearance. However, the advantages of deep appearance representations in Random Finite Set based filters, such as the GM-PHD filter, have not been explored which works online and run fast enough to be suitable for real-time applications, especially in efficiently integrating deep appearance representations at both the update step of the filter and the target labeling stage.

In this work, we propose an online multi-object visual tracker based on the GM-PHD filter using a tracking-by-detection approach for real-time applications which not only runs in real-time but also addresses track management (target birth, death and labeling), false positives and miss-detections jointly. We also learn discriminative deep appearance representations of targets using identification network (IdNet) on large-scale person re-identification data sets. We formulate how to combine (fuse) spatio-temporal and visual similarities obtained from bounding boxes of objects and their CNN appearance features, respectively, to construct a cost to be minimized (similarity maximized) by the Hungarian algorithm to label each target. We also formulate an augmented likelihood using CNN appearance features and motion information and then integrate into the update step of the GM-PHD filter. After this association step, additional unassigned tracks prediction step is used to overcome the miss-detection susceptibility of the GM-PHD filter caused by occlusion. Furthermore, we use the deeply learned CNN appearance representations as a person re-identification method to re-identify lost objects for consistently labeling them. To the best of our knowledge, nobody has adopted this approach.

The main contributions of this paper are as follows:
\begin{enumerate}
  \item We use the GM-PHD filter with the deeply learned CNN features to develop a real-time tracker to track multiple targets in video sequences acquired under varying environmental conditions and targets density.
  \item We formulate how to integrate spatio-temporal and visual similarities obtained from bounding boxes of objects and their CNN appearance features at both the update step of the GM-PHD filter and the target labeling stage.
  \item We include additional unassigned tracks predictions after the association step to overcome the miss-detection susceptibility of the GM-PHD filter.
  \item We use the deeply learned CNN appearance representations as a person re-identification method to re-identify lost objects for consistently labeling them.
  \item We make extensive evaluations on Multiple Object Tracking 2016 (MOT16), 2017 (MOT17) and HiEve benchmark data sets using the public detections provided in the benchmark's test sets.
\end{enumerate}

We presented a preliminary idea of this work in~\cite{Nat19}. In this work, we make more elaborate descriptions of our algorithm. In addition, we change from joint-input Siamese network (StackNet) to identification network (IdNet) to learn the deep appearance representations of targets on large-scale person re-identification data sets as this IdNet allows us to extract features from an object once in each frame in the tracking process which speeds up the tracker significantly. We also formulate an augmented likelihood using deep appearance features and motion information and then integrate into the update step of the GM-PHD filter. Furthermore, we include an additional add-on prediction step for predicting unassigned tracks after the association step to handle miss-detections caused by occlusion.

The paper is organized as follows. We discuss the related work in Section~\ref{Sec:RelatedWork}. In Section~\ref{Sec:proposedAlgorithm}, our proposed algorithm is explained in detail including its all components, and Section~\ref{Sec:ParameterValues} provides some important parameter values in the GM-PHD filter implementation. The experimental results are analyzed and compared in Section~\ref{Sec:ExperimentalResults}. The main conclusions and suggestions for future work are summarized in Section~\ref{Sec:Conclusions}.

\section{Related Work} \label{Sec:RelatedWork}

Numerous multi-target tracking algorithms have been introduced in the literature~\cite{LuoZhaKim14}\cite{VoMalBarCorOsbMahVo15}\cite{PatPanLil18}. Traditionally, multi-target trackers have been developed by finding associations between targets and observations mainly using JPDAF~\cite{BarWilTia11} and MHT~\cite{BarWilTia11,KimLiCip15}. However, these approaches have faced challenges not only in the uncertainty caused by data association but also in algorithmic complexity that increases exponentially with the number of targets and measurements.

Recently, a unified framework which directly extends single to multiple target tracking by representing multi-target states and observations as RFS was developed by Mahler~\cite{Mah03} which not only addresses the problem of increasing complexity, but also estimates the states and cardinality of an unknown and time varying number of targets in the scene by allowing for target birth, death, clutter (false alarms), and missing detections. It propagates the first-order moment of the multi-target posterior, called intensity or the PHD~\cite{VoMa06}, rather than the full multi-target posterior. This approach is flexible, for instance, it has been used to find the detection proposal with the maximum weight as the target position estimate for tracking a target of interest in dense environments by removing the other detection proposals as clutter~\cite{BaiBhoWal18}\cite{Nat18}. Furthermore, the standard PHD filter was extended to develop a novel N-type PHD filter ($N \geq 2$) for tracking multiple targets of different types in the same scene~\cite{NatAnd19}\cite{BaiWal19}. However, this approach does not naturally include target identity in the framework because of the indistinguishability assumption of the point process; additional mechanism is necessary for labeling each target. Recently, labeled RFS for multi-target tracking was introduced in~\cite{VoVoPhu14}\cite{VoVoHoa17}\cite{Kim17}, however, its computational complexity is high. In general, the RFS-based filters are susceptible to miss-detection even though they are robust to clutter.

The two common implementation schemes of the PHD filter are the Gaussian mixture (GM-PHD filter~\cite{VoMa06}) and Sequential Monte Carlo (SMC-PHD or particle PHD filter~\cite{VoSinDou05}). Though the PHD filter is the most widely adopted RFS-based filter in computer vision due to its computational efficiency (it has a linear complexity with number of targets and observations), it is weak in handling miss-detection. This is because the PHD filter is designed originally for radar tracking applications where the number of miss-detections is very low as opposed to the visual tracking applications where significant number of miss-detections occur due to occlusion. The work in~\cite{SonYooYow19} tried to alleviate the miss-detection problem of the PHD filter using occlusion group management strategy. However, it introduces an energy minimization problem for obtaining an optimal hypothesis which poses more computation though the author(s) speeded it up using C++ implementation. In this work, we overcome not only the miss-detection problem but also the labeling of targets in each frame for real-time visual tracking applications.

Incorporating deep appearance information into multi-target tracking algorithms improves the tracking performance as demonstrated in works such as~\cite{KimLiCip15,KimLiReh18,Kim17,FuSngNaq18,Baisa2019,JinChaYan20,JinFanJoh18}. In fact, object detection, appearance modeling, motion modeling and filtering are the main components of visual tracking. The way these components are integrated and how they are modeled makes difference. For instance, multi-output regularized least squares (MORLS) framework has been used to learn appearance models online and are integrated into a tree-based track-oriented MHT (TO-MHT) in~\cite{KimLiCip15}. The same author has trained a bilinear long short-term memory (LSTM) on both motion and appearance and has incorporated it into the MHT for gating in~\cite{KimLiReh18}. A tracklet siamese network with constrained clustering has been proposed in~\cite{JinFanJoh18}. These trackers are, however, computationally demanding and operate offline. A joint detection, feature extraction and tracking with an end-to-end training has been proposed in~\cite{JinChaYan20}. Appearance models of objects are also learned in the same fashion as in~\cite{KimLiCip15} to integrate into a generalized labeled multi-Bernoulli (GLMB) filter~\cite{Kim17}. Deep discriminative correlation filters have also been learned and integrated into the PHD filter in~\cite{FuSngNaq18,ZeyFedJon19}. Though the latter three trackers~(\cite{Kim17,FuSngNaq18,Baisa2019}) work online, they are computationally demanding to be applied to time-critical real-time applications.

The two well-known CNN structures are verification and identification models~\cite{ZhengZY16}. In general, Siamese network, a kind of verification network (similarity metric), is the most widely used network for developing multi-target tracking methods~\cite{LeaCanSch16}\cite{TanAndAnd17}\cite{Nat19}\cite{YooKimYo19}. As discussed in~\cite{LeaCanSch16}, the Siamese topology has three types: those combined at cost function, in-network and StackNet. The StackNet which has been used in offline tracking~\cite{LeaCanSch16}\cite{TanAndAnd17} and online tracking~\cite{Nat19}\cite{YooKimYo19} methods outperforms the other types of Siamese topologies. This StackNet can also be referred to as joint-input network~\cite{YooKimYo19}. This network takes two concatenated image patches along the channel dimension and infers their similarity. The last fully-connected layer of the StackNet models a 2-way classification problem (the same and different identities) i.e. given a pair of images, the StackNet produces the probability of the pair being the same or different identity by a forward pass. This means in multi-target tracking applications, all pair of tracks and detections (estimates in our case) need to be paired and given as input to the StackNet to get their probability of similarity in each video frame. This leads to a high complexity as demonstrated in~\cite{LeaCanSch16}\cite{TanAndAnd17}\cite{Nat19}\cite{YooKimYo19} which limits the trackers' applications for real-time scenarios. We observed this in our preliminary work~\cite{Nat19}, thus, we change the StackNet to identification network (IdNet) compensating for the performance by training the IdNet on large-scale person re-identification data sets; the StackNet generally outperforms the IdNet~\cite{TanAndAnd17}. Using this IdNet, appearance features are extracted once in each video frame from detections (or output estimates from the GM-PHD filter) and are copied to the assigned tracks after the data association step. This speeds up the online tracker very significantly when compared to using the StackNet.

In addition to learning the discriminative deep appearance representations to incorporate an augmented likelihood into the GM-PHD filter, solve tracks-to-estimates associations and lost tracks re-identifications, we also include additional add-on unsigned tracks prediction after the association step to over-come the miss-detection problem of the GM-PHD filter due to occlusion. To date, no work has incorporated these all important components not only to improve the multi-target tracking performance but also to speed it up to the level of real-time, as is the case in our work.

\section{The Proposed Algorithm} \label{Sec:proposedAlgorithm}

\begin{figure*}[htbp] 
\begin{center}
  \includegraphics[width=1.0\linewidth]{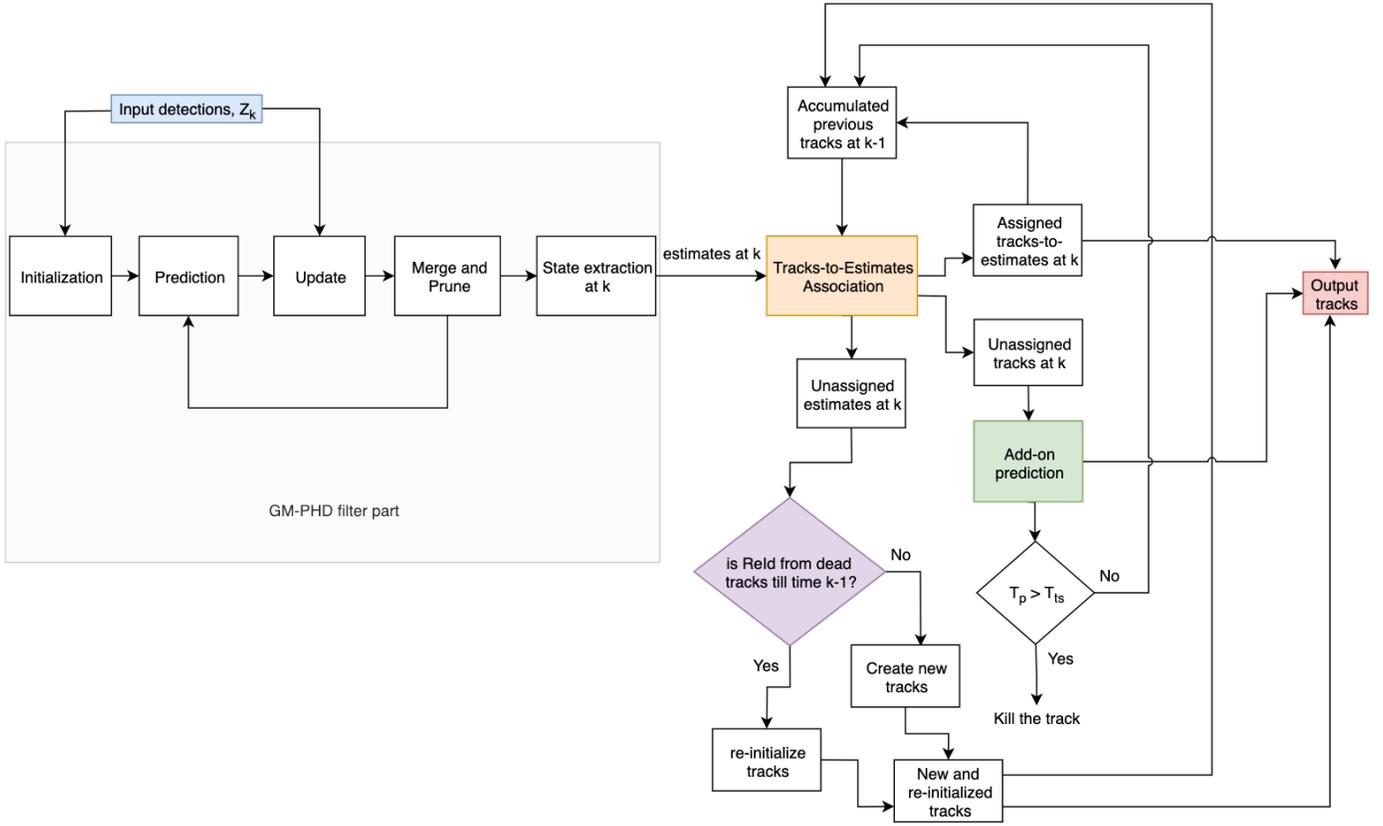} \\
\end{center}
   \caption{Block diagram of the proposed multi-target visual tracking pipeline using the GM-PHD filter, visual-spatio-temporal information for tracks-to-estimates association, lost tracks re-identification using deep visual similarity and additional add-on unassigned tracks prediction.}
\label{fig:MOTdiagram}
\end{figure*}

The block diagram of our proposed multi-target tracking algorithm is given in Fig.~\ref{fig:MOTdiagram}. Our proposed online tracker consists of four components: 1) target states estimation using the GM-PHD filter, 2) tracks-to-estimates associations using the Hungarian algorithm, 3) add-on unassigned tracks prediction to alleviate miss-detections, and 4) lost tracks re-identification for tracks re-initialization. All of these four components are explained in details as follows.

\noindent

\subsection{The GM-PHD Filter} \label{Sec:GMPHD-Filter}

The Gaussian mixture implementation of the standard PHD (GM-PHD) filter~\cite{VoMa06} is a closed-form solution of the PHD filter that assumes a linear Gaussian system. It has two steps: prediction and update. Before stating these two steps, certain assumptions are needed: 1) each target follows a linear Gaussian model:

\begin{equation}
    f_{k|k-1}(x|\zeta) =  \mathcal{N}(x;F_{k-1}\zeta, Q_{k-1})
\label{eqn:linearState1}
\end{equation}
\noindent
\begin{equation}
    g_{k}(z_d|x) =  \mathcal{N}(z_d;H_{k} x, R_{k})
\label{eqn:linearObservation1}
\end{equation}
\noindent where $f_{k|k-1}(.|\zeta)$ is the single target state transition probability density at time k given the previous state $\zeta$ and $g_{k}(z_d|x)$ is the single-target (single-measurement association) likelihood function due to motion which defines the probability that $z_d$ is generated (observed) conditioned on state $x$. $\mathcal{N}(.;m, P)$ denotes a Gaussian density with mean $m$ and covariance $P$; $F_{k-1}$ and $H_k$ are the state transition and measurement matrices, respectively. $Q_{k-1}$ and $R_k$ are the covariance matrices of the process and the measurement noises, respectively. The measurement noise covariance $R_k$ can be measured off-line from sample measurements i.e. from ground truth and detection of training data~\cite{WelBis06} as it indicates detection performance.
2) A current measurement driven birth intensity inspired by but not identical to~\cite{RisClaVoVo12} is introduced at each time step, removing the need for the prior knowledge (specification of birth intensities) or a random model, with a non-informative zero initial velocity. The intensity of the spontaneous birth RFS is a Gaussian mixture of the form

\begin{equation}
\begin{split}
     \gamma_{k}(x)  =  \sum_{v = 1}^{V_{\gamma,k}} w_{\gamma,k}^{(v)}\mathcal{N}(x; m_{\gamma,k}^{(v)}, P_{\gamma,k}^{(v)})
\label{eqn:PHDbirthassumptionLCMHT}
\end{split}
\end{equation}
\noindent where $V_{\gamma,k}$ is the number of birth Gaussian components, $w_{\gamma,k}^{(v)}$ is the weight accompanying the Gaussian component $v$, $m_{\gamma,k}^{(v)}$ is the current measurement and zero initial velocity used as mean, and $P_{\gamma,k}^{(v)}$ is birth covariance for Gaussian component $v$.

3) The survival and detection probabilities are independent of the target state: $p_{S,k}(x_k) = p_{S,k}$ and $p_{D,k}(x_k) = p_{D,k}$.

\textbf{Adaptive birth:} We use adaptive measurement-driven approach for birth of targets. Each detection $z_k \in Z_k$ is associated with detection confidence score $s_k \in [0, 1]$. We use more confident (strong) detections based on their score for birth of targets as they are more likely to represent a potential target. Confident detections used for birth of targets will be $Z_{b,k} = \{z_{b,k}: s_k \geq s_t\} \subseteq Z_k$ where $s_t$ is a detection score threshold. In fact, $s_t$ governs the relationship between the number of false positives (clutter) and miss-detections (false negatives). Increasing the value of $s_t$ gives more miss-detections and less false positives, and vice versa. The initial birth weight $w_{\gamma,k}^{(v)}$ in Eq.~(\ref{eqn:PHDbirthassumptionLCMHT}) is also weighted by $s_k$ to give high probability for more confident detections for birth of targets i.e. $w_{\gamma,k}^{(v)} = s_kw_{\gamma,k}^{(v)}$. However, all measurements $Z_k$ are used for the update step.

\textbf{Prediction:} It is assumed that the posterior intensity at time $k-1$ is a Gaussian mixture of the form

\begin{equation}
\begin{split}
     \mathcal{D}_{k-1}(x)  = \mathcal{D}_{k-1|k-1}(x) = \sum_{v = 1}^{V_{k-1}} w_{k-1}^{(v)}\mathcal{N}(x; m_{k-1}^{(v)}, P_{k-1}^{(v)}),
\label{eqn:PHDposterior1k-1}
\end{split}
\end{equation}

\noindent where $V_{k-1}$ is the number of Gaussian components of $\mathcal{D}_{k-1}(x)$ and it equals to the number of Gaussian components after pruning and merging at the previous iteration. Under these assumptions, the predicted intensity at time $k$ is given by

\begin{equation}
    \mathcal{D}_{k|k-1}(x) = \mathcal{D}_{S,k|k-1}(x) + \gamma_{k}(x),
\label{eqn:PHDpredictionI1}
\end{equation}
\noindent where

\begin{equation}
\begin{array} {lll}  \mathcal{D}_{S,k|k-1}(x) =& p_{S,k} \sum_{v = 1}^{V_{k-1}} w_{k-1}^{(v)}\mathcal{N}(x;  m_{S,k|k-1}^{(v)},P_{S,k|k-1}^{(v)}), \nonumber
\end{array}
\label{eqn:PHDpredictionSurvival1}
\end{equation}
\noindent
\begin{equation}
 m_{S,k|k-1}^{(v)} = F_{k-1} m_{k-1}^{(v)},  \nonumber
\label{eqn:PHDpredictionSurvivalMean1}
\end{equation}
\noindent
\begin{equation}
 P_{S,k|k-1}^{(v)} = Q_{k-1} + F_{k-1} P_{k-1}^{(v)} F^T_{k-1},  \nonumber
\label{eqn:PHDpredictionSurvivalCov1}
\end{equation}
\noindent where $\gamma_k(x)$ is given by Eq.~(\ref{eqn:PHDbirthassumptionLCMHT}).

Since $\mathcal{D}_{S,k|k-1}(x)$ and $\gamma_k(x)$ are Gaussian mixtures, $ \mathcal{D}_{k|k-1}(x)$ can be expressed as a Gaussian mixture of the form

\begin{equation}
\begin{split}
     \mathcal{D}_{k|k-1}(x)  =  \sum_{v = 1}^{V_{k|k-1}} w_{k|k-1}^{(v)}\mathcal{N}(x; m_{k|k-1}^{(v)},P_{k|k-1}^{(v)}),
\label{eqn:PHDpredictionki}
\end{split}
\end{equation}
\noindent where $w_{k|k-1}^{(v)}$ is the weight accompanying the predicted Gaussian component $v$, and $V_{k|k-1}$ is the number of predicted Gaussian components and it equals to the number of born targets and the number of persistent (surviving) components. The number of persistent components is actually the number of Gaussian components after pruning and merging at the previous iteration. 

\textbf{Update with Augmented Likelihood:} In this step, we modify the likelihood due to motion given in Eq.~(\ref{eqn:linearObservation1}) by including appearance likelihood to formulate an augmented likelihood to incorporate into the update step of the GM-PHD filter. In this case, let $z = (z_d, z_w)\in Z_k$ be an ordered pair of a detection vector $z_d$ and an appearance feature vector $z_w$ which is extracted from the detected object region i.e. a region enclosed by a detection vector $z_d$ (object bounding box represented as a vector). Assuming a paired observation $z \in Z_k$ is an ordered pair conditionally independent observations, the augmented likelihood can be given by

\begin{equation}
g_k(z|x) = g_k(z_d|x) g_k(z_w|x),
\label{eq:AugmentedLikelihood}
\end{equation}
\noindent where the detection vector likelihood $g_k(z_d|x)$ is provided in Eq.(\ref{eqn:linearObservation1}) whereas the appearance feature likelihood function $g_k(z_w|x)$ will be discussed in Section~\ref{Sec:DataAssociation} i.e. given in Eq.~(\ref{eq:AppearanceLikelihood}).

Thus, the posterior intensity (updated PHD) at time $k$  is also a Gaussian mixture and is given by incorporating the augmented likelihood as follows
\begin{equation}
\begin{split}
     \mathcal{D}_{k|k}(x)  =  (1 - p_{D,k})\mathcal{D}_{k|k-1}(x) + \sum_{z\in Z_k} \mathcal{D}_{D,k}(x;z),
\label{eqn:PHDupdatekiLCMHT}
\end{split}
\end{equation}
\noindent where
\begin{equation}
\begin{split}
     \mathcal{D}_{D,k}(x;z)  =  \sum_{v = 1}^{V_{k|k-1}} w_{k}^{(v)}(z) \mathcal{N}(x; m_{k|k}^{(v)}(z_d), P_{k|k}^{(v)}), \nonumber
\label{eqn:PHDupdateDetki}
\end{split}
\end{equation}
\noindent
\begin{equation}
\begin{split}
     w^{(v)}_{k}(z)  =  \frac{p_{D,k} w^{(v)}_{k|k-1} q^{(v)}_{k}(z)}{c_{s_{k}}(z_d) +  p_{D,k} \sum_{l = 1}^{V_{k|k-1}} w^{(l)}_{k|k-1} q^{(l)}_{k}(z)}, \nonumber
\label{eqn:PHDupdatewwightki}
\end{split}
\end{equation}
\noindent
\begin{equation}
\begin{split}
     q^{(v)}_{k}(z)  = g_k(z_w|x) \mathcal{N}(z_d; H_k m_{k|k-1}^{(v)}, R_k + H_k P_{k|k-1}^{(v)} H^T_k), \nonumber
\label{eqn:PHDupdateqki}
\end{split}
\end{equation}
\noindent
\begin{equation}
\begin{split}
     m^{(v)}_{k|k}(z_d)  =  m^{(v)}_{k|k-1} + K^{(v)}_k (z_d - H_k m_{k|k-1}^{(v)}), \nonumber
\label{eqn:PHDupdatemki}
\end{split}
\end{equation}
\noindent
\begin{equation}
\begin{split}
     P^{(v)}_{k|k}  =  [I -  K^{(v)}_k H_k] P_{k|k-1}^{(v)}, \nonumber
\label{eqn:PHDupdatepki}
\end{split}
\end{equation}
\noindent
\begin{equation}
\begin{split}
     K^{(v)}_k  =  P_{k|k-1}^{(v)} H^T_k [ H_k P_{k|k-1}^{(v)} H^T_k + R_k]^{-1} \nonumber
\label{eqn:PHDupdateKki}
\end{split}
\end{equation}
\noindent The clutter intensity due to the scene, $c_{s_k}(z_d)$, in Eq.~(\ref{eqn:PHDupdatekiLCMHT}) is given by
\begin{equation}
    c_{s_k}(z_d) = \lambda_t c(z_d) = \lambda_{c} A c(z_d),
\label{eqn:Clutterscenei}
\end{equation}
\noindent where $c(.)$ is the uniform density over the surveillance region $A$, and $\lambda_{c}$ is the average number of clutter returns per unit volume i.e. $\lambda_t = \lambda_{c}A$. 

After update, weak Gaussian components with weight $w_k^{(v)} < T = 10^{-5}$ are pruned, and Gaussian components with Mahalanobis distance less than $U = 4$ pixels from each other are merged. We limit the number of Gaussian components to the maximum of $V_{k-1}$, $M_k$ (number of measurements) and a sample from a Poisson distribution with $V_{k-1}$ as mean. These Gaussian components are chosen after sorting them based on their weights; the strong ones are selected. This increases the speed of our tracker. These pruned and merged Gaussian components are predicted as existing (persistent) targets in the next iteration. Finally, Gaussian components of the pruned and merged intensity with means corresponding to weights greater than 0.5 as a threshold are selected as multi-target state estimates (we use the pruned and merged intensity rather than the posterior intensity as it gives better results).

\subsection{Data Association}   \label{Sec:DataAssociation}

The GM-PHD filter distinguishes between true and false targets, however, this does not distinguish between two different targets, so an additional step is necessary to identify different targets between consecutive frames. We use both the spatio-temporal and visual similarities between the track boxes and estimated object states (filtered output boxes) in frames $k-1$ and $k$, respectively, to label each object across frames.

\subsubsection{Spatio-temporal information}

The spatio-temporal information is computed using track boxes and filtered output boxes in consecutive frames. Let $b^t_{i,k-1}$ be the $i^{th}$ track's box and $b^e_{j,k}$ be the $j^{th}$ estimate's (GM-PHD filter's filtered output) box at frame k. Their spatio-temporal similarity is calculated using Euclidean distance $D_k(b^t_{i,k-1},b^e_{j,k})$ between their centers. We use Euclidean distance rather than Jaccard distance (1 - Intersection-over-Union) as it gives slightly better result. The spatio-temporal (motion or distance) relation has been commonly used, in different forms, in many multi-object tracking works~\cite{MatPoiCav16}\cite{TanAndSch17}\cite{YuLiLi16}. The normalized Euclidean distance $D_{n,k}(b^t_{i,k-1},b^e_{j,k})$ between the centers of the bounding boxes $b^t_{i,k-1}$ and $b^e_{j,k}$ is given by

\begin{equation}
\begin{array} {lll}
     D_{n,k}(b^t_{i,k-1},b^e_{j,k}) =\\ \sqrt{\bigg(\frac{b^t_{i,x,k-1} - b^e_{j,x,k}}{W}\bigg)^2 + \bigg(\frac{b^t_{i,y,k-1} - b^e_{j,y,k}}{H}\bigg)^2},
\end{array}
\label{eqn:Dist}
\end{equation}
\noindent where $(b^t_{i,x,k-1}, b^t_{i,y,k-1})$ and $(b^e_{j,x,k}, b^e_{j,y,k})$ are the center locations of their corresponding bounding boxes at frames $k-1$ and $k$, respectively. $W$ and $H$ are the width and height of a video frame which are used for the Euclidean distance normalization.

\subsubsection{Deep Appearance Representations Learning}

Visual cues are very crucial for associating tracks with detections (in our case current filtered outputs or estimated states) for robust online multi-object tracking. In this work, we propose an identification CNN network (IdNet) for computing visual affinities between image patches cropped at bounding box locations. We treated this task as a multi-class recognition problem to learn a discriminative CNN embedding. We adopted the ResNet~\cite{KaiXiaSha15} as the network structure (ResNet50) by replacing the topmost layer (fully connected layer) to output confidence for each of the person identities in the training data set (changing from 1000 classes to 6654 classes in our case). We also add another fully-connected layer to compress 2048-dimensional embedding to 512-dimensional embedding just after the last adaptive average pooling layer along with one-dimensional batch normalization and a dropout with a rate of 0.5 for reducing a possible over-fitting. The rest of the ResNet50 architecture remains the same except removing the final downsample by changing a stride value from 2 to 1 which increases performance. We use a transfer learning approach rather than training the network from scratch i.e. the pre-trained weights on the ImageNet data set~\cite{ILSVRC15} consisting of 1000 classes is first loaded before modifying the network. For the modified or newly added layers, we initialize the weights and bias terms. For the first-fully-connected layer, we use the Kaiming's initialization method using a normal distribution for initializing the weights. We use normal distribution with mean 1.0 and standard deviation (std) of 0.02 for initializing the weights of the one-dimensional batch normalization. Normal distribution with mean 0 and std of 0.001 is used for the initialization of the weights of the last fully-connected layer (classifier). Constant zero value is used for initializing the bias terms of all the modified or newly added layers (two fully-connected layers and one-dimensional batch normalization).

\textbf{Data preparation: }To learn discriminative deep appearance representations, we collected our training data set from numerous sources. First, we utilize publicly available person re-identification data sets including Market1501 data set~\cite{ZheSheTia15} (736 identities from 751 as we restrict the number of images per identity to at least 4), DukeMTMC data set~\cite{RisSolZou16} (702 identities), CUHK03 data set~\cite{LiZhaXia14} (1367 identities), LPW data set~\cite{sonLenLiu17} (1974 identities), and MSMT data set~\cite{LonShiWen17} (1141 identities). In addition to these person re-identification data sets, we also collected training data from publicly available tracking data sets such as MOT15~\cite{MOT15} and MOT16/17~\cite{MOT16} training data sets (MOT16 and MOT17 have the same training data set though MOT17 is claimed to have more accurate ground truth and is used in our experiment). From all these tracking training data sets of MOT15 (TUD-Stadmitte, TUD-Campus, PETS09-S2L1, ETH-Bahnhof and ETH-Sunnyday) and MOT16/17 (5 sequences), we produce about 521 person identities. We also produce about 213 identities from TownCentre data set~\cite{BenRei11}. This helps the network more to adapt to the MOT benchmark test sequences as well as the network can learn the inter-frame variations. In total, we collected about 6,654 person identities from all these data sets to train our IdNet. One image is randomly chosen from each identity in this training set to be used as a validation image. We resize all the training images to $284\times142$ and then subtract the mean image from all the images which is computed from all the training images. During training, we augment the training samples by randomly cropping all the images to $256\times128$ and then randomly flipping horizontally to increase more variation and thus reduce possible over-fitting. Other data augmentation techniques, such as color jittering and random erasing with a random erasing probability of 0.5, are also used in the training phase. We also use a random order of images by reshuffling the data set.

\textbf{Training: }We train the IdNet using cross-entropy loss (softmax loss) and mini-batch Stochastic Gradient Descent (SGD) with momentum. The mini-batch size, the momentum  and the weight-decay factor for L2 regularization are set to 32, 0.9 and $5e^{-4}$, respectively. We also enabled the Nesterov momentum. We trained our model on a NVIDIA GeForce GTX 1050 GPU for 100 epochs, after which it generally converges, using PyTorch deep learning framework~\cite{PyTorch19}. We initialize the learning rate to $2 \times 10^{-3}$ for all parameters except for the last modified or newly added two fully-connected layers' parameters for which we multiplied the learning rate by 10 to give a higher learning rate. We also use a learning rate scheduler i.e. we decay the learning rate by a factor of 0.1 every 40 epochs (1 epoch is one sweep over all the training data).

\textbf{Testing: }We use 2 video sequences of MOT16/17 training data set (02 and 09) for testing our trained IdNet. For this testing set, we produce about 66 person identities. We randomly sample about 800 positive pairs (the same identities) and 3200 negative pairs (different identities) from ground truth of MOT16/17-02 and MOT16/17-09 training dataset. We use this larger ratio of negative pairs to mimic the positive/negative distribution during tracking. We use verification accuracy as an evaluation metric. Given an input image, we extract a 512-dimensional embedding using our trained model by first removing the last fully-connected layer (the classifier). Thus, given a pair of images, we compute the cosine distance (using Eq.~(\ref{eq:CosineD})) between their extracted deep appearance feature vectors. If the computed cosine-distance of positive pairs are greater than or equal to 0.75, they are assumed as correctly classified pairs. Similarly, if the computed cosine-distance of negative pairs are less than 0.75, they are assumed as correctly classified pairs. Accordingly, the IdNet trained on large-scale data sets (6,654 identities) gives about 97.5\% accuracy.

\subsubsection{Tracks-to-Estimates Association}

Here we use visual-spatio-temporaal information, fusion of both visual and spatio-temporal information, to associate tracks to the estimated (filtered output) boxes.

The visual similarity $V_{s,k}(b^t_{i,k-1},b^e_{j,k})$ between the track's box $b^t_{i,k-1}$ and estimate's (filtered output) box $b^e_{j,k}$ at frame k is computed using the cosine distance $C_d(\mathbf{z}^i_w, \mathbf{z}^j_w)$ between appearance feature vectors $\mathbf{z}^i_w$ and $\mathbf{z}^j_w$ which are extracted from the track's box $b^t_{i,k-1}$ and filtered output (detection) box $b^e_{j,k}$, respectively. Thus, this visual similarity (cosine distance) is given using the dot product and magnitude (norm) of the appearance feature vectors as

\begin{equation}
V_{s,k}(b^t_{i,k-1},b^e_{j,k}) = C_d(\mathbf{z}^i_w, \mathbf{z}^j_w) =  \frac{\mathbf{z}^i_w \cdot \mathbf{z}^j_w}{\| \mathbf{z}^i_w \| \|\mathbf{z}^j_w\|},
\label{eq:CosineD}
\end{equation}
\noindent We consider the mean of temporal track features till frame $k-1$, temporal track features are aggregated to be represented as a single embedding, when computing the cosine distance between track features and estimated state features. Though it is possible to extract deep appearance features from estimates, in this work we extract deep appearance features from detections and propagate throughout the GM-PHD filtering process to keep the ones corresponding to the estimates and then to the tracks. In this way, it is possible to use the extracted deep appearance features at both the update step of the GM-PHD filter and the target labeling stage. Thus, the appearance feature likelihood function $g_k(z_w|x)$ in Eq.~(\ref{eq:AugmentedLikelihood}) is given by

\begin{equation}
g_k(z_w|x) =  \frac{\exp(C_d(\mathbf{z}^i_w, \mathbf{z}^j_w))}{\exp(C_d(\mathbf{z}^i_w, \mathbf{z}^j_w)) + \exp(-C_d(\mathbf{z}^i_w, \mathbf{z}^j_w))},
\label{eq:AppearanceLikelihood}
\end{equation}
\noindent

The visual-spatio-temporal similarity is utilized to construct a bipartite graph between the tracks and estimates. We use the Munkres's variant of the Hungarian algorithm~\cite{FraJea71} to determine the optimal associations in case an estimate (filtered output) box is tried to be associated with multiple tracks using the following overall association cost

\begin{equation}
     \mathbf{C}_k  =  (1 - \eta) \mathbf{D}_{n,k} + \eta \mathbf{V}_{d,k},
\label{eqn:OverallSimilarity}
\end{equation}
\noindent
where $\mathbf{V}_{d,k} = \mathbf{1} - \mathbf{V}_{s,k}$ is the visual difference used as a cost where each of its element $V_{d,k} \in [0, 2]$, $\mathbf{D}_{n,k}$ is a matrix of the normalized Euclidean distances where each element $D_{n,k} \in [0, 1]$, and $\eta$ is the weight balancing the two costs. $\mathbf{C}_k \in \mathbb{R}^{N\times M}$, $\mathbf{D}_{n,k} \in \mathbb{R}^{N\times M}$ and $\mathbf{V}_{d,k} \in \mathbb{R}^{N\times M}$ are matrices where $N$ and $M$ are the number of tracks and estimates (filtered outputs) at time $k$; $\mathbf{1}$ is a matrix of $1's$ of the same dimension as $\mathbf{V}_{s,k} \in \mathbb{R}^{N\times M}$. Spatio-temporal relation gives useful information for tracks-to-estimates association of targets that are in close proximity, however, its importance starts to decrease as targets become (temporally) far apart. In contrast, visual similarity obtained from CNN allows long-range association as it is robust to large temporal and spatial distance. These combination of spatio-temporal and visual information helps to solve target ambiguities which may occur due to either targets motion or their visual content as well as allows long-range association of targets.

The outputs of the Hungarian algorithm are assigned tracks-to-estimates, unassigned tracks and unassigned estimates as shown in Fig.~\ref{fig:MOTdiagram}. The tracks-to-estimates association is confirmed if the cost $C_k(b^t_i,b^e_j)$ is lower than the cost threshold $C_{ts} = 0.4$. The associated estimates (filtered outputs) boxes are appended to their corresponding tracks to generate longer ones up to time k. The unassigned tracks are predicted using the add-on prediction step or killed accordingly as discussed in Section~\ref{Sec:EstimatesPrediction}. The unassigned estimates either create new tracks or perform re-identification from the lost (dead) tracks to re-initialize the tracks as discussed in Section~\ref{Sec:ReId}.

\subsection{Unassigned tracks Prediction}  \label{Sec:EstimatesPrediction}

We keep state transition matrix ($F_{k-1}$), process noise covariance ($Q_{k-1}$) and the covariance matrices ($P_{k-1}$) from the update step of the GM-PHD filter for the unassigned tracks obtained after the Hungarian algorithm-based tracks-to-estimates association step. We, therefore, predict each of the unassigned track $X^t_{k-1}$ using its state transition matrix while also updating its covariance matrix $P^t_{k-1}$ for a period of $T_p$ number of predictions (frames) as follows (Eq.~\ref{eqn:AddOnPred}).

\begin{equation}
\begin{split}
       X^t_k =& F_{k-1}X^t_{k-1}  \\
       P^t_k =& Q_{k-1} + F_{k-1} P^t_{k-1} F^T_{k-1}
\label{eqn:AddOnPred}
\end{split}
\end{equation}
\noindent where $X^t_k$ and $P^t_k$ are the updated unassigned track $t$'s state (location) and covariance matrix at frame $k$, respectively.

The effect of this additional add-on prediction step versus the number of predictions ($T_p$) is analyzed in Table~\ref{tbl:MOT16TrainingTp}. We kill the track if the number of performed predictions is greater than the number of predictions threshold ($T_{ts}$). This killed track can be considered for re-identification in the upcoming frames. In our experiment, we choose $T_{ts} = 3$ as this gives better Multiple Object Tracking Accuracy (MOTA) value as shown in Table~\ref{tbl:MOT16TrainingTp} and Fig.~\ref{fig:MOTAvsTp}. Detailed investigation of this add-on prediction on the performance of our online tracker is given in experimental results Section~\ref{Sec:AblationStudy}.

\subsection{Re-identification for Tracking} \label{Sec:ReId}

Person re-identification in the context of multi-target tracking is very challenging due to occlusions (inter-object and background), cluttered background and inaccurate bounding box localization. Inter-object occlusions are very challenging in video sequences containing dense targets, hence, object detectors may miss some targets in some consecutive frames. Re-identification of lost targets due to miss-detections is crucial to keep track of identity of each target.

The tracks-to-estimates association using the Hungarian algorithm given in Section~\ref{Sec:DataAssociation} can also provide unassigned estimates. If a past track is not associated to any estimated box at frame k, the tracked target might be occluded or temporally missed by the object detector. If an estimated object box is not associated to any track, it is used for initializing a new track if it is not created earlier by checking it within the last $m = 1:k-1$ frames from lost or dead tracks using visual similarity $V_{s,k}$ for re-identification. We use a visual similarity threshold of $V^s_{ts} = 0.6$ for the re-identification of targets i.e re-identification occurs if the visual similarity (cosine distance) is greater than $V^s_{ts} = 0.6$. If multiple dead tracks are matched to the unassigned estimate, the one with the maximum similarity score is confirmed. Re-identification using the visual similarity along with combining the visual similarity with the spatio-temporal information to construct the cost for labeling of targets and incorporating the augmented likelihood into the update step of the GM-PHD filter has increased the performance of our online tracker as shown in Table~\ref{tbl:MOT16TrainingTp} and Fig.~\ref{fig:MOTA1601}. An independent analysis of each component is also given in experimental results Section~\ref{Sec:AblationStudy}.

\section{Parameter Values in the GM-PHD Filter Implementation} \label{Sec:ParameterValues}

Our state vector includes the centroid positions, velocities, width and height of the bounding boxes,  i.e. $x_k = [p_{cx,xk}, p_{cy,xk}, \dot{p}_{x,xk}, \dot{p}_{y,xk}, w_{xk}, h_{xk}]^T$.
Similarly, the measurement is the noisy version of the target area in the image plane approximated with a $w$ x $h$ rectangle centered at $(p_{cx,xk}, p_{cy,xk})$ i.e. $z_k = [p_{cx,zk}, p_{cy,zk}, w_{zk}, h_{zk}]^T$.

We set the survival probability $p_{S} = 0.99 $, and we assume the linear Gaussian dynamic model of Eq.~(\ref{eqn:linearState1}) with matrices taking into account the box width and height at the given scale.

\[ F_{k-1} = \left[ \begin{array}{ccc}
           I_2 & \Delta I_2 & 0_2 \\
           0_2 & I_2 & 0_2 \\
           0_2 & 0_2 & I_2
           \end{array} \right], \]
													
\begin{equation} Q_{k-1} = \sigma_v^2 \left [ \begin{array}{ccc}
    \frac{\Delta^4}{4}I_2 & \frac{\Delta^3}{2}I_2  & 0_2\\
    \frac{\Delta^3}{2}I_2 & \Delta^2 I_2 & 0_2 \\
    0_2 & 0_2 &  I_2
    \end{array} \right],
\label{eqn:PHDstateTransitionMatrixVideo}
\end{equation}
\noindent
where $F$ and $Q$ denote the state transition matrix and process noise covariance, respectively; $I_n$ and $0_n$ denote the \textit{n} x \textit{n} identity and zero matrices, respectively, and $\Delta = 1$ second is the sampling period defined by the time between frames.  $\sigma_v = 5$ pixels$/s^2$ is the standard deviation of the process noise.

Similarly, the measurement follows the observation model of Eq.~(\ref{eqn:linearObservation1}) with matrices taking into account the box width and height,

\[H_k = \left[ \begin{array}{ccc}
           I_2 & 0_2 & 0_2 \\
           0_2 & 0_2 & I_2
           \end{array} \right], \]

\begin{equation} R_k = \sigma_r^2 \left [ \begin{array}{cc}
     I_2 & 0_2 \\
     0_2 & I_2
    \end{array} \right],
\label{eqn:eqn:PHDobservationMatrixVideo}
\end{equation}
\noindent where $H_k$ and $R_k$ denote the observation matrix and the observation noise covariance, respectively, and $\sigma_r = 6$ pixels is the measurement standard deviation. The probability of detection is assumed to be constant across the state space and through time and is set to a value of $p_D = 0.95$. The false positives are independently and identically distributed (i.i.d), and the number of false positives per frame is Poisson-distributed with mean $\lambda_t  = 10$ (false alarm rate of $\lambda_c \approx 4.8 \times 10^{-6}$; dividing the mean $\lambda_t$ by frame resolution $A$, refer to Eq.~(\ref{eqn:Clutterscenei})).

Nothing is known about the appearing targets before the first observation. The distribution after the observation is determined by the current measurement and zero initial velocity used as a mean of the Gaussian distribution and using a predetermined initial covariance given in Eq.~(\ref{eqn:PHDbirthCovariance}) for birthing of targets.

\begin{equation}
 P_{\gamma,k}  = diag([100, 100, 25, 25, 20, 20]).
\label{eqn:PHDbirthCovariance}
\end{equation}
\noindent

The birth weight $w_{\gamma,k}$ that any potential observation represents an appearing target in Eq.~(\ref{eqn:PHDbirthassumptionLCMHT}), detection score threshold $s_t$ and whether using detection score $s_k$ along with (multiplied by) $w_{\gamma,k}$ depends on the application as they govern the relationship between false positives and miss-detections i.e. they are hyper-parameters that require tuning. In our evaluations, we find $w_{\gamma,k} = 0.1$, $s_t = 0.0$ and without using $s_k$ along with $w_{\gamma,k}$ gives better MOTA value at the expense of increased false positives. The influence of $s_k$ and $s_t$ partly depends on the value of $w_{\gamma,k}$. Furthermore, after evaluating on $\eta \in \{0.0, 0.4, 0.65, 0.85, 1.0\}$ (in Eq.~(\ref{eqn:OverallSimilarity})), we set it to 0.65 as this gives better result. The implementation parameters and their values are summarized in Table~\ref{tbl:parameters}.

\begin{table*}[htbp]
\begin{center}
\begin{tabular}{|l|c|c|c|c|c|c|c|c|c|c|c|c|c|r|}
\hline
Parameters & $\eta$ &  $s_t$ & $w_{\gamma,k}$ & $\sigma_r$ & $\sigma_v$ & $p_D$ & $p_S$ & $\lambda_t$ & $U$ & $T$ & $m$ & $T_{ts}$ & $C_{ts}$ & $V^s_{ts}$\\
\hline
Values  & 0.65 & 0.0 & 0.1 & 6 pixels & 5 pixels$/s^2$ & 0.95 & 0.99 & 10 & 4 pixels & $10^{-5}$ & 1:k-1 frames & 3 & 0.4 & 0.6 \\
\hline
\end{tabular}
\end{center}
\caption{Implementation values of the parameters used in our evaluations on both MOT16 and MOT17 Benchmark data sets~\cite{MOT16}; for both ablation study (Table~\ref{tbl:MOT16TrainingTp}) and comparison with other trackers (Table~\ref{tbl:MOT16} and Table~\ref{tbl:MOT17}).}
\label{tbl:parameters}
\end{table*}
\noindent

\section{Experimental Results} \label{Sec:ExperimentalResults}

In this Section, we discuss experimental settings, ablation Study on MOT Benchmark training set and evaluations on MOT Benchmark test sets in detail.

\subsection{Experimental Settings}

The experimental settings for the proposed online tracker such as tracking data sets, evaluation metrics and implementation details are presented as follows.

\textbf{Tracking Datasets: }We make an extensive evaluations of our proposed online tracker using both MOT16 and MOT17 benchmark data sets~\cite{MOT16} which are captured on unconstrained environments using both static and moving cameras. These data sets consist of 7 training sequences on which we make ablation study as given in Section~\ref{Sec:AblationStudy} (Table~\ref{tbl:MOT16TrainingTp}) and 7 testing sequences on which we evaluate and compare our proposed online tracker with other trackers as shown in Table~\ref{tbl:MOT16} and Table~\ref{tbl:MOT17}. We use the \textit{public detections} provided by the MOT benchmark with a non-maximum suppression (NMS) of 0.3 for DPM detector~\cite{FelGirMcA10} (for both MOT16 and MOT17) and 0.5 for FRCNN~\cite{ShaKaiRos15} and SDP~\cite{YanChoLin16} detectors (for MOT17).

\textbf{Evaluation Metrics: }We use numerous evaluation metrics including the CLEAR metrics~\cite{BerSti08}, the identity preservation measure~\cite{RisSolZou16} and the set of track quality measures~\cite{LiHua09} which are presented as follows:
\begin{itemize}
  \item Multiple Object Tracking Accuracy (MOTA): A summary of overall tracking accuracy in terms of false positives, false negatives and identity switches, which gives a measure of the tracker's performance at detecting objects as well as keeping track of their trajectories.
  \item Multiple Object Tracking Precision (MOTP):  A summary of overall tracking precision in terms of bounding box overlap between ground-truth and tracked location, which shows the ability of the tracker to estimate precise object positions.
  \item Identification F1 (IDF1) score~\cite{RisSolZou16}: The quantitative measure obtained by dividing the number of correctly identified detections by the mean of the number of ground truth and detections.
  \item Mostly Tracked targets (MT): Percentage of mostly tracked targets (a target is tracked for at least 80\% of its life span regardless of maintaining its identity) to the total number of ground truth trajectories.
  \item Mostly Lost targets (ML): Percentage of mostly lost targets (a target is tracked for less than 20\% of its life span) to the total number of ground truth trajectories.
  \item False Positives (FP): Number of false detections.
  \item False Alarms per Frame (FAF): This can also be referred to as false positives per image (FPPI) which measures false positive ratio.
  \item False Negatives (FN): Number of miss-detections.
  \item Identity Switches (IDSw): Number of times the given identity of a ground-truth track changes.
  \item Fragmented trajectories (Frag): Number of times a track is interrupted (compared to ground truth trajectory) due to miss-detection.
\end{itemize}
\noindent
True positives are detections which have at least 50\% overlap with their corresponding ground truth bounding boxes. For more detailed description of each metric, please refer to~\cite{MOT16}. The implementation of these all multi-object evaluation metrics are given in MATLAB\footnote{https://motchallenge.net/devkit} and in Python\footnote{https://github.com/cheind/py-motmetrics}.

\textbf{Implementation Details: }Our proposed tracking algorithm is implemented in Python on a i7 2.80 GHz core processor with 8 GB RAM. We use the PyTorch deep learning framework~\cite{PyTorch19} for CNN feature extraction where its forward propagation computation is transferred to a NVIDIA GeForce GTX 1050 GPU, and our tracker runs at about 31.16 frames per second (fps). The forward propagation for feature extraction step is relatively the main computational load of our tracking algorithm. However, it is much significantly faster than our preliminary work in~\cite{Nat19} (3.5 fps) since appearance features are extracted once from detections in each frame and then copied to the associated tracks rather than concatenating both track and estimate patches along the channel dimension and extracting the features from all tracks and estimates in every frame~\cite{Nat19}.


\subsection{Ablation Study on MOT16 Benchmark Training Set} \label{Sec:AblationStudy}

\begin{figure}
\centering
\begin{minipage}[c]{0.48\textwidth}
\centering
    \includegraphics[width=1.0\linewidth]{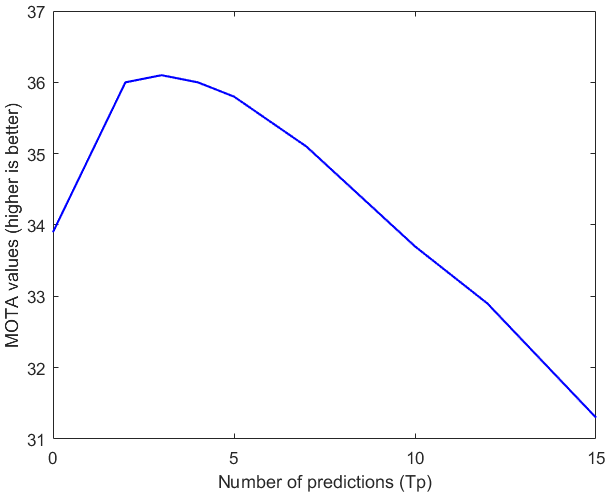}
    \caption{MOTA values of the proposed GMPHD-ReId tracker when the number of predictions ($T_p$) is varied on the \textbf{MOT16}~\cite{MOT16} benchmark training set. Maximum MOTA value is obtained at $T_p = T_{ts} = 3$.}
    \label{fig:MOTAvsTp}
\end{minipage}
\end{figure}

We investigate the contributions of the different components of our proposed online tracker, GMPHD-ReId, on the MOT16~\cite{MOT16} benchmark training set using public detections. These different components include motion information (Mot), appearance information for augmented likelihood and labeling (App), re-identification (ReId) and add-on unassigned tracks predictions (AddOnPr). First, we evaluate using only the motion information (Mot) as shown in Table~\ref{tbl:MOT16TrainingTp}. Second, we include appearance information (App) and re-identification (ReId) in addition to the motion information to see the effect of the learned discriminative deep appearance representations on the tracking performance. Third, we include the additional add-on prediction (AddOnPr) on top of the motion information, appearance information and re-identification, particularly by varying the number of predictions ($T_p$) as shown in Table~\ref{tbl:MOT16TrainingTp} using numerous tracking evaluation metrics. The graphical plot the MOTA values in Table~\ref{tbl:MOT16TrainingTp} versus the number of predictions ($T_p$) is shown in Fig.~\ref{fig:MOTAvsTp}.

Accordingly, using only the motion information provides MOTA value of 32.1 and IDF1 of 26.1 as shown in Table~\ref{tbl:MOT16TrainingTp}. Including the deeply learned appearance information for augmented likelihood, data association and re-identification increases the MOTA and IDF1 to 33.9 and 44.5, respectively. This is an increase by 5.61\% for MOTA and by 70.50\% for IDF1. We also investigate the influence of the additional add-on prediction (AddOnPr) step by varying the number of predictions $T_p$ from 0 (no AddOnPr) to 15. The maximum MOTA value is obtained at $T_p = 3$ as shown in Table~\ref{tbl:MOT16TrainingTp} and Fig.~\ref{fig:MOTAvsTp}. Thus, including the additional add-on prediction with $T_p = 3$ in the our proposed online tracker increases the MOTA and IDF1 from 33.9 to 36.1 and from 44.5 to 46.9, respectively. This means an increase of 6.49\% and 5.39\% for MOTA and IDF1, respectively, is obtained using a very simple additional add-on unassigned tracks prediction. Thus, each component of the our proposed online tracker has an effect of increasing tracking performance.

\begin{table*}[htbp]
\begin{center}
\begin{tabular}{|l|c|c|c|c|c|c|c|c|c|r|}
\hline
Type  & MOTA$\uparrow$ & IDF1$\uparrow$ & MOTP$\uparrow$ & FAF$\downarrow$& MT (\%)$\uparrow$ & ML (\%)$\downarrow$ & FP$\downarrow$ & FN$\downarrow$ & IDS$\downarrow$ & Frag$\downarrow$ \\
\hline
Mot Only & 32.1 & 26.1	& \textbf{77.7}	& \textbf{0.33} & 4.00 &	26.50 &	\textbf{1738} & 70796 &	2427 &	2549  \\
Mot + App + ReId + 0 AddOnPr & 33.9	& 44.5 & 77.6 & 0.34 &	4.10 &	26.40 &	1744 & 70794 &	468 &	2550 \\
Mot + App + ReId + 2 AddOnPr &  36.0 & 46.8 & 77.0 & 0.64 &	5.80 &	24.20 &	3400 & 66938 & \textbf{375} & 1419  \\
Mot + App + ReId + 3 AddOnPr &  \textbf{36.1} & 46.9 & 76.8 & 0.79 &	5.90 &	23.70 &	4193 &	65976 &	\textbf{375} & 1294  \\
Mot + App + ReId + 4 AddOnPr &  36.0 & 46.9 & 76.7 & 0.95 & 6.30 & 23.50 & 5032 &	65240 &	378	& 1236	\\
Mot + App + ReId + 5 AddOnPr & 	35.8 & 47.5 & 76.5 & 1.11 & 6.40 &	22.90 &	5878 &	64611 &	386	& 1196  \\
Mot + App + ReId + 7 AddOnPr & 	35.1 & 47.6 & 76.3 & 1.43 & 6.90 &	22.60 &	7580 &	63650 &	377 & 1160  \\
Mot + App + ReId + 10 AddOnPr & 33.7 & 47.3 &	76.1 &	1.90	& 7.00 & 22.30 & 10101 & 62704 & 380 & 1117 \\
Mot + App + ReId + 12 AddOnpr &	32.9 & \textbf{48.0} &	76.0 & 2.18 &	7.20 &	22.20 &	11604 &	62147 &	385 & 1099	\\
Mot + App + ReId + 15 AddOnpr &	31.3 & 47.7 & 75.9	& 2.62 & \textbf{7.50} & \textbf{21.90} &	13951 &	\textbf{61522} &	382	& \textbf{1091} \\
\hline
\end{tabular}
\end{center}
\caption{Tracking performance evaluation results on the \textbf{MOT16}~\cite{MOT16} benchmark training set using public detections in terms different GMPHD-ReId components: motion information (Mot), appearance information (App), re-identification (ReId) and add-on unassigned tracks predictions (AddOnPr). Evaluation measures with ($\uparrow$) show that higher is better, and with ($\downarrow$) denote lower is better. In this experiment, using motion, appearance, reid and add-on unassigned tracks predictions for 3 consecutive frames gives the best MOTA value.}
\label{tbl:MOT16TrainingTp}
\end{table*}

\begin{figure*}[htbp] 
  \begin{center}
  {\label{fig:GMPHDFrame04} \includegraphics[height=0.145\textwidth]{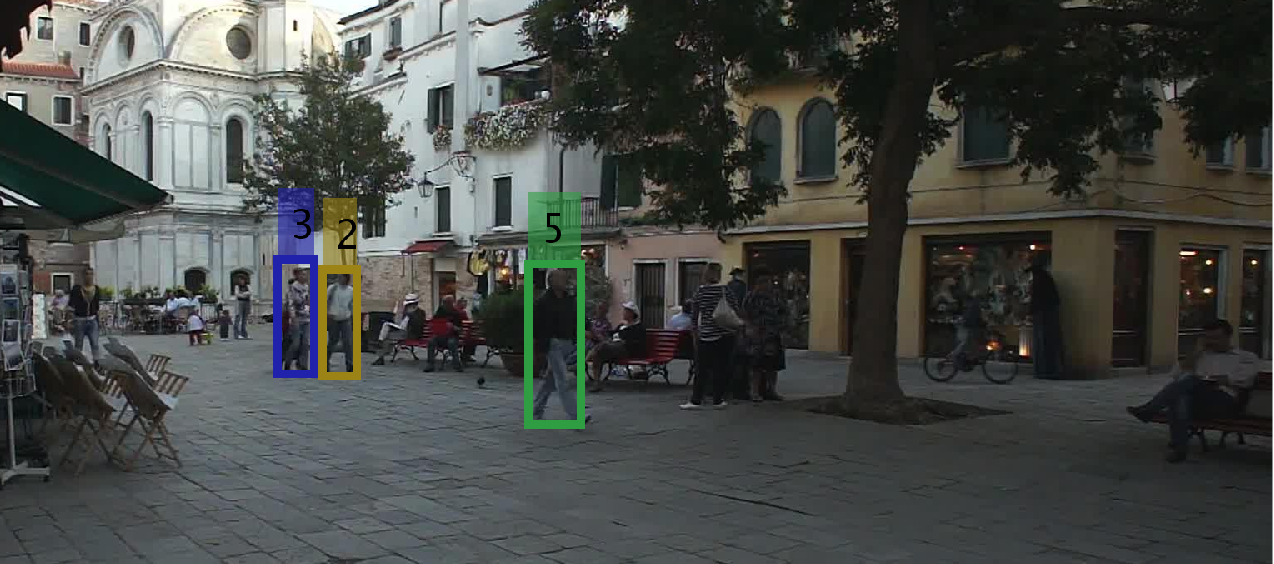}} 
  {\label{fig:GMPHDFrame17} \includegraphics[height=0.145\textwidth]{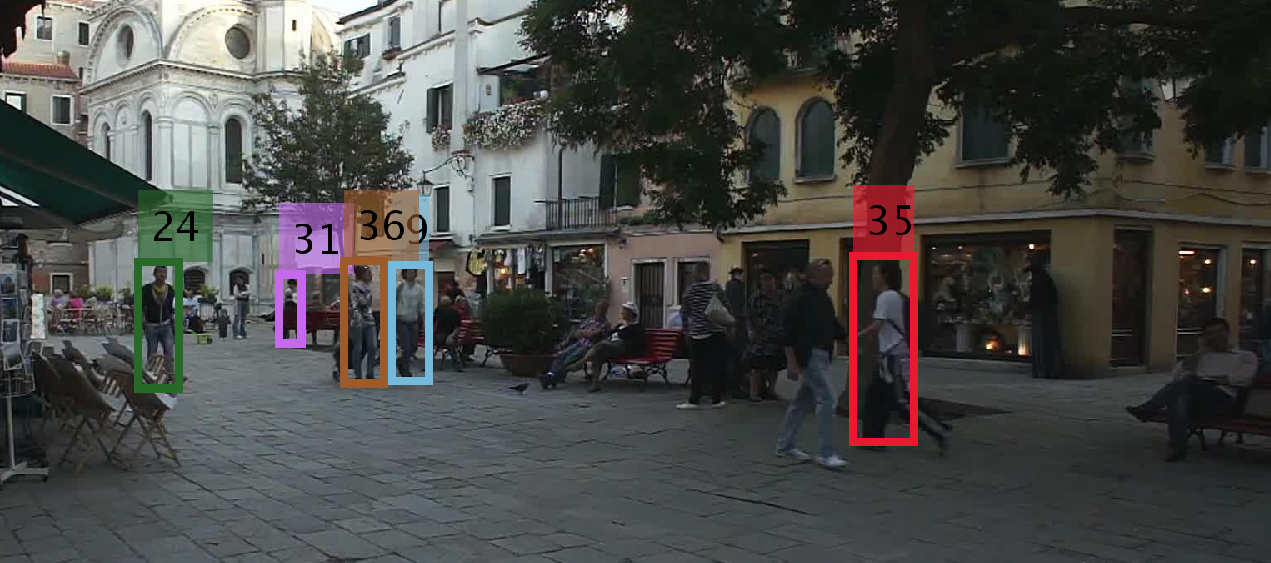}} 
  {\label{fig:GMPHDFrame36} \includegraphics[height=0.145\textwidth]{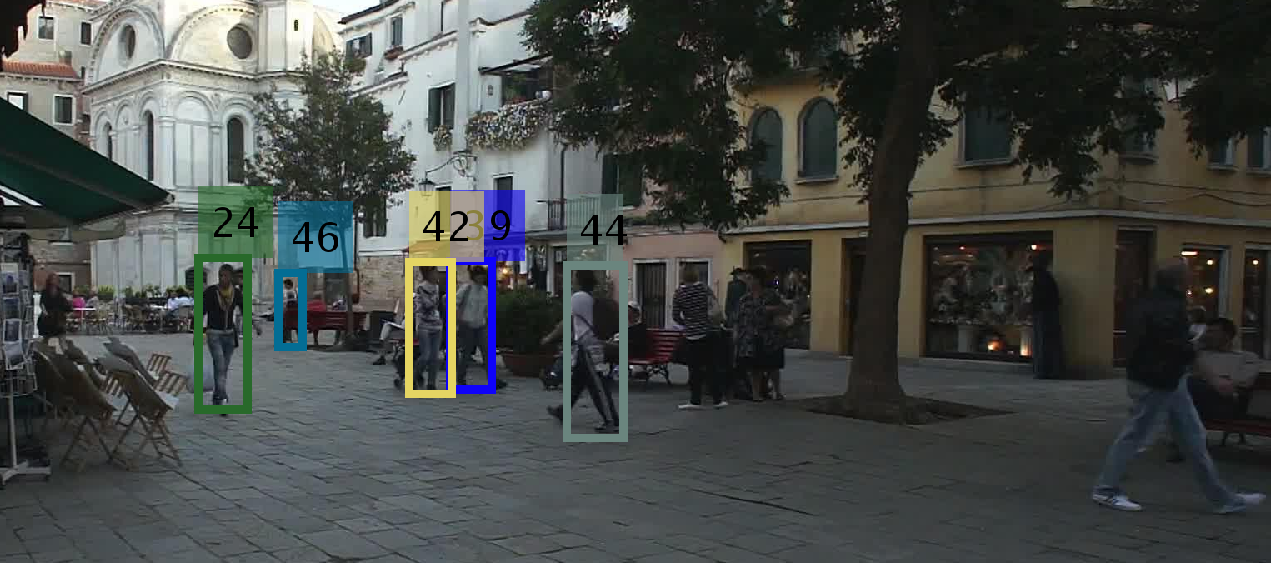}}  \\  
  {\label{fig:GMPHDINTFrame04} \includegraphics[height=0.145\textwidth]{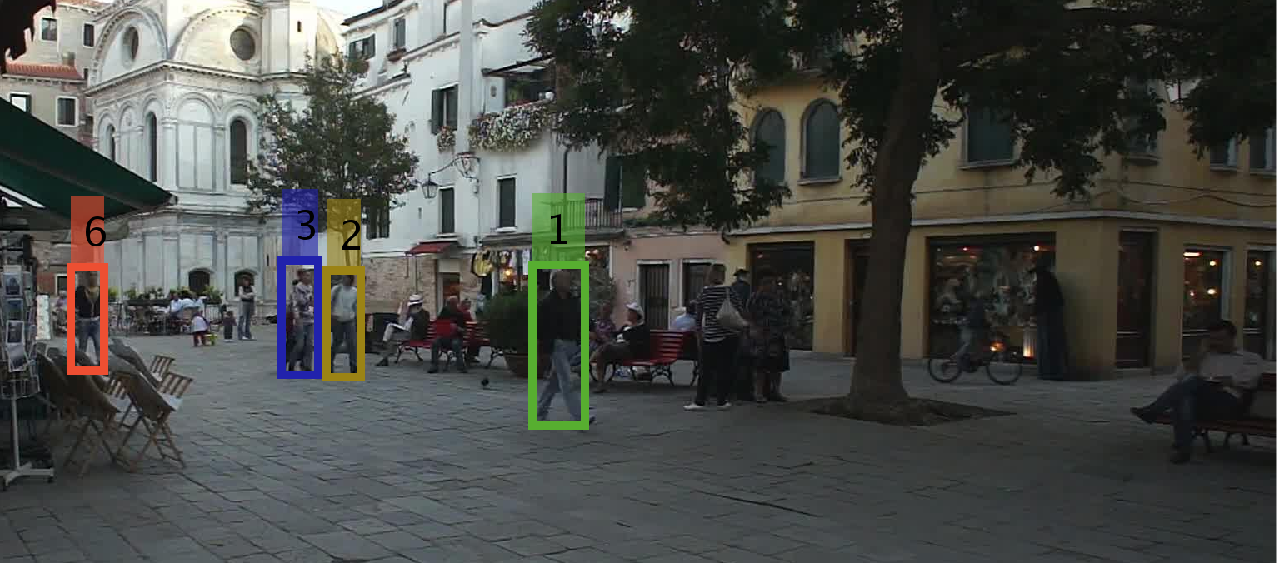}}
  {\label{fig:GMPHDINTFrame04} \includegraphics[height=0.145\textwidth]{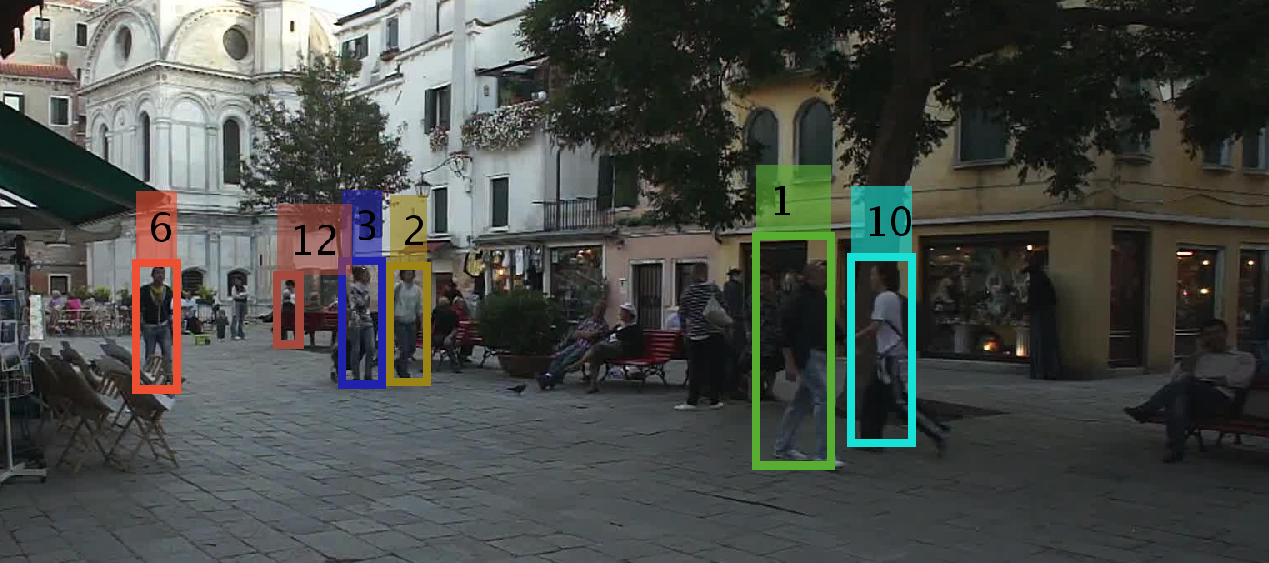}} 
  {\label{fig:GMPHDINTFrame04} \includegraphics[height=0.145\textwidth]{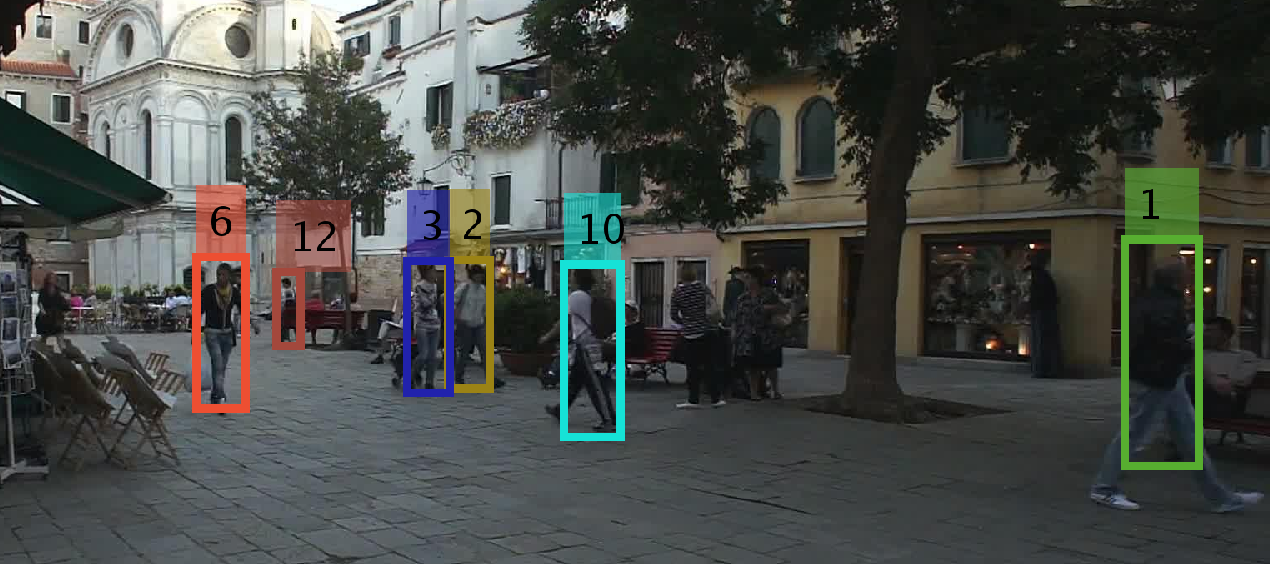}}  
  \end{center}
   \caption{Sample results on 3 frames of MOT16-01 data set for our proposed online tracker with motion information only (top row for frames 11, 75 and 125 from left to right) and with appearance, re-identification and add-on prediction (bottom row for frames 11, 75 and 125 from left to right). Bounding boxes represent the tracking results with their color-coded identities; small numbers are also shown on top of each bounding box for better clarity.}
  \label{fig:MOTA1601}
\end{figure*}
\noindent

\subsection{Evaluations on MOT Benchmark Test Sets} 

After validating our proposed tracker, GMPHD-ReId, on the MOT16 Benchmark training set with the add-on prediction with $T_p = 3$ in Section~\ref{Sec:AblationStudy}, we compare it against state-of-the-art online and offline tracking methods as shown in Table~\ref{tbl:MOT16} and Table~\ref{tbl:MOT17}. Accordingly, quantitative evaluations of our proposed method with other trackers is compared in Table~\ref{tbl:MOT16} on MOT16 benchmark data set. The Table shows that our algorithm outperforms both online and offline trackers listed in the table in many of the evaluation metrics. When compared to the online trackers, our proposed online tracker outperforms all the others in MOTA, IDF1, MT, ML and FN. The number of identity switches (IDSw) is also significantly lower than many of the online trackers. Our proposed online tracker outperforms not only many of the online trackers but also several offline trackers in terms of several evaluation metrics. In terms of IDF1, our proposed online tracker performs better than all of the trackers, both online and offline, listed in the table. Our online tracker also runs faster than many of both online and offline trackers, at about 31.6 fps.

Our online tracker also gives promising results on MOT17 benchmark data set as is quantitatively shown in Table~\ref{tbl:MOT17}. It outperforms all other online trackers in the table in all IDF1, MT, ML and Hz (speed) measures. The number of FN, IDSw and Frag is also significantly lower than many of the online trackers in the table. Even though MOTDT~\cite{CheAiZhu18}, FAMNet~\cite{ChuLin19} and GMPHDOGM17~\cite{SonYooYow19} outperform our tracker in terms of MOTA, our proposed tracker outperforms all these trackers in terms of many other evaluation metrics such as IDF1, MT, ML and speed (Hz) as can been from the Table~\ref{tbl:MOT17}. For instance, in terms of speed, our tracker is 51.33 times faster than the FAMNet (30.8 Hz vs 0.6 Hz) and 1.68 times faster than the MOTDT (30.8 Hz vs 18.3 Hz). In addition to the online trackers, our proposed online tracker outperforms many of the offline trackers listed in the table. Our proposed online tracker outperforms almost all the trackers in the table, both online and offline including the well-known offline trackers such as MHT-DAM~\cite{KimLiCip15} and MHT-bLSTM~\cite{KimLiReh18}, in terms of IDF1 and ML, while running in real-time.

The most important to notice here is that the comparison of our algorithm to the GM-PHD-HDA~\cite{SonJeo16} (GMPHD-SHA for MOT17). These both trackers use the GM-PHD filter but with different approaches for labeling of targets from frame to frame. While our tracker uses the Hungarian algorithm for labeling of targets by postprocessing the output of the filter using a combination of spatio-temporal and visual similarities along with visual similarity for re-identification, the GM-PHD-HDA uses the approach in~\cite{PanClaVo09} at the prediction stage by also including appearance features for re-identification to label targets. In addition to the GM-PHD-HDA tracker, our proposed tracker outperforms the other GM-PHD filter-based trackers such as GMPHD-KCF~\cite{KutBosEis17}, GM-PHD-N1T (GMPHD-N1Tr)~\cite{BaiWal19} and GM-PHD-DAL (GMPHD-DAL)~\cite{Nat19} (our preliminary work) as shown in Tables~\ref{tbl:MOT16} and~\ref{tbl:MOT17} in almost all of the evaluation metrics. As shown in Table~\ref{tbl:MOT16}, our proposed online tracker is not only faster than our preliminary work~\cite{Nat19} (3.5 fps versus 31.6 fps), but also increases the tracking accuracy from MOTA value of 35.1 to 40.4 (by about 15.10\%). When compared to GMPHDOGM17~\cite{SonYooYow19}, our tracker uses only one step association and is faster even in Python implementation (GMPHDOGM17 is implemented in C++ i.e. the approach has more computational and algorithmic complexity than ours).

The qualitative comparison of our proposed tracker (GMPHD-ReId) and our tracker without appearance information and additional unassigned tracks prediction is given in Fig.~\ref{fig:MOTA1601} for frames 11, 75 and 125. Due to the detection failures, some labels of targets are not consistent for our tracker without appearance information and additional unassigned tracks prediction (top row), for instance, labels 2 and 3 in frame 11 are changed to labels 9 and 36, respectively, in frame 75. Similarly labels 31, 35 and 36 in frame 75 are changed to labels 46, 44 and 42, respectively, in frame 125. However, the labels of the targets are consistent when using the GMPHD-ReId tracker (bottom row). The effect of the additional unassigned tracks prediction is also clearly visible overcoming occlusion problems. For instance, a person with label 5 in frame 11 is missed in 75 and 125 frames when using our tracker without appearance information and additional unassigned tracks prediction (top row), however, this same person with label 1 in frame 11 is tracked in both 75 and 125 frames when using our proposed online tracker which combines all the components together (bottom row): motion information, deep appearance information for augmented likelihood, data association and re-identification, and additional add-on unassigned tracks prediction.

In our evaluations, the association cost constructed using only visual similarity CNN gives better result than using only spatio-temporal relation, however, their combination using Eq.~(\ref{eqn:OverallSimilarity}) gives better result than each of them. Furthermore, weighted summation of the costs according to Eq.~(\ref{eqn:OverallSimilarity}) gives slightly better result than the Hadamard product (element-wise multiplication) of the two costs.

\begin{table*}
\begin{center}
\begin{tabular}{|l|c|c|c|c|c|c|c|c|c|c|r|}
\hline
Tracker & Mode & MOTA$\uparrow$ & MOTP$\uparrow$ & IDF1$\uparrow$ & MT (\%)$\uparrow$ & ML (\%)$\downarrow$ & FP$\downarrow$ & FN$\downarrow$ & IDSw$\downarrow$ & Frag$\downarrow$ & Hz $\uparrow$ \\
\hline
MHT-DAM~\cite{KimLiCip15} & offline & \color{red}{\textbf{45.8}} & \color{blue}{\textbf{76.3}}	& \color{blue}{\textbf{46.1}} & \color{red}{\textbf{16.2}} &  \color{red}{\textbf{43.2}} &	6,412 & \color{red}{\textbf{91,758}}	& \color{blue}{\textbf{590}} & 781 & 0.8 \\
MHT-bLSTM6~\cite{KimLiReh18} & offline & \color{blue}{\textbf{42.1}} & 75.9 & \color{red}{\textbf{47.8}} & \color{blue}{\textbf{14.9}} & \color{blue}{\textbf{44.4}} & 11,637 & \color{blue}{\textbf{93,172}} & 753 & 1,156 & 1.8 \\
CEM~\cite{MilRotSch14} & offline & 33.2 & 75.8 & N/A & 7.8 &	54.4 &	6,837 &	114,322 & 642 &	\color{blue}{\textbf{731}} & 0.3\\
DP-NMS~\cite{PirRamFow11} & offline & 32.2 & \color{red}{\textbf{76.4}} & 31.2 & 5.4 & 62.1 &	\color{red}{\textbf{1,123}} &	121,579 & 972 &	944 & \color{blue}{\textbf{5.9}} \\
SMOT~\cite{DicCamSzn13} & offline & 29.7 & 75.2 & N/A & 5.3 & 47.7 &	17,426 & 107,552 & 3,108 & 4,483  & 0.2\\
JPDF-m~\cite{RezMilZha15} & offline & 26.2 & \color{blue}{\textbf{76.3}} & N/A & 4.1 &	67.5 &	\color{blue}{\textbf{3,689}} &	130,549 & \color{red}{\textbf{365}} & \color{red}{\textbf{638}} & \color{red}{\textbf{22.2}} \\
\hline
GM-PHD-HDA~\cite{SonJeo16} & \textbf{online} & 30.5 & 75.4 & 33.4 &	4.6 & 59.7 & 5,169 & 120,970 & \color{red}{\textbf{539}} & \color{red}{\textbf{731}} & 13.6\\
GM-PHD-N1T~\cite{BaiWal19} & \textbf{online} & 33.3 & \color{red}{\textbf{76.8}} & 25.5 & 5.5 & 56.0 & \color{red}{\textbf{1,750}} & 116,452 & 3,499 & 3,594 & 9.9 \\
HISP-T~\cite{Nat18} & \textbf{online} & 35.9 & 76.1 & 28.9 & 7.8 & 50.1 & 6,406 & 107,905 & 2,592 & 2,299  & 4.8\\
OVBT~\cite{BanBaAla16} & \textbf{online} & 38.4 & 75.4 & 37.8 & 7.5 & \color{blue}{\textbf{47.3}} & 11,517 & \color{blue}{\textbf{99,463}} & 1,321 & 2,140 & 0.3 \\

EAMTT-pub~\cite{MatPoiCav16} & \textbf{online} & \color{blue}{\textbf{38.8}}  & 75.1 & \color{blue}{\textbf{42.4}} & \color{blue}{\textbf{7.9}} & 49.1 & 8,114 & 102,452 & 965 & 1,657 & 11.8\\
JCmin-MOT~\cite{BorJeo17} & \textbf{online} & 36.7 & 75.9 & 36.2 & 7.5 & 54.4 &  2,936	& 111,890 &	\color{blue}{\textbf{667}}	& \color{blue}{\textbf{831}} & \color{blue}{\textbf{14.8}} \\
GM-PHD-DAL~\cite{Nat19} & \textbf{online} & 35.1 & \color{blue}{\textbf{76.6}} & 26.6 & 7.0 & 51.4 & \color{blue}{\textbf{2,350}} & 111,886 & 4,047 & 5,338 & 3.5\\
\textbf{GMPHD-ReId (ours)} & \textbf{online} & \color{red}{\textbf{40.4}} & 75.2 & \color{red}{\textbf{50.1}} & \color{red}{\textbf{11.5}} & \color{red}{\textbf{43.1}} & 6,569 & \color{red}{\textbf{101,251}} & 789 & 2,519 & \color{red}{\textbf{31.6}} \\
\hline
\end{tabular}
\end{center}
\caption{Tracking performance of representative trackers developed using both online and offline methods. All trackers are evaluated on the test data set of the \textbf{MOT16}~\cite{MOT16} benchmark using public detections. The first and second highest values are highlighted by $\color{red}{\textbf{red}}$ and $\color{blue}{\textbf{blue}}$, respectively (for both online and offline trackers). Evaluation measures with ($\uparrow$) show that higher is better, and with ($\downarrow$) denote lower is better. N/A shows not available.}
\label{tbl:MOT16}
\end{table*}

\begin{table*}[htbp]
\begin{center}
\begin{tabular}{|l|c|c|c|c|c|c|c|c|c|c|r|}
\hline
Tracker & Mode & MOTA$\uparrow$ & MOTP$\uparrow$ & IDF1$\uparrow$ & MT (\%)$\uparrow$ & ML (\%)$\downarrow$ & FP$\downarrow$ & FN$\downarrow$ & IDSw$\downarrow$ & Frag$\downarrow$  & Hz $\uparrow$ \\
\hline
MHT-DAM~\cite{KimLiCip15} & offline & 50.7 & \color{red}{\textbf{77.5}} & 47.2 & 20.8 & \color{blue}{\textbf{36.9}} & 22,875 & 252,889 &	2,314 & 2,865 & 0.9 \\
MHT-bLSTM~\cite{KimLiReh18} & offline & 47.5	& \color{red}{\textbf{77.5}} & 51.9 & 18.2 &	41.7 &	25,981 & 268,042 & 2,069 & 3,124 & 1.9\\
IOU17~\cite{BocEisSik17} & offline & 45.5 & \color{blue}{\textbf{76.9}} & 39.4 & 15.7 & 40.5 & 19,993 & 281,643 & 5,988 & 7,404 & \color{red}{\textbf{1,522.9}}\\

TPM~\cite{JinTaoWei20} & offline & \color{blue}{\textbf{54.2}} & N/A & 52.6 &	\color{blue}{\textbf{22.8}} &	37.5 &	\color{blue}{\textbf{13,739}}	& \color{blue}{\textbf{242,730}} &  1,824 &	\color{blue}{\textbf{2,472}} &  0.8 \\

MPNTrack~\cite{GuiLau20} & offline & \color{red}{\textbf{58.8}} & N/A & \color{red}{\textbf{61.7}} & \color{red}{\textbf{28.8}} & \color{red}{\textbf{33.5}} & 17,413 & \color{red}{\textbf{213,594}} & \color{red}{\textbf{1,185}} & \color{red}{\textbf{2,265}} & 6.5 \\	

SAS-MOT17~\cite{MakFua19} & offline & 44.2 & 76.4 & \color{blue}{\textbf{57.2}} & 16.1 & 	44.3 &	29,473 & 283,611 & \color{blue}{\textbf{1,529}} & 2,644 &  4.8\\
DP-NMS~\cite{PirRamFow11} & offline & 43.7 & \color{blue}{\textbf{76.9}} & N/A & 12.6 & 46.5 & \color{red}{\textbf{10,048}} & 302,728 & 4,942 & 5,342 & \color{blue}{\textbf{137.7}} \\
\hline
EAMTT~\cite{MatPoiCav16} & \textbf{online} & 42.6 & 76.0 & 41.8 & 12.7 & 42.7 & 30,711 & 288,474 & 4,488 & 5,720 & 12.0\\
FPSN~\cite{LeeKim19} & \textbf{online} & 44.9 & 76.6 & 48.4 & 16.5 & 35.8 & 33,757 & 269,952 & 7,136 & 14,491 & 10.1 \\
OTCD-1~\cite{LiuWuLi19} & \textbf{online} & 44.9 & \color{blue}{\textbf{77.4}} & 42.3 & 14.0 &	44.2 &	\color{blue}{\textbf{16,280}}	& 291,136 & 3,573 & 5,444 & 5.5\\
MOTDT~\cite{CheAiZhu18} & \textbf{online} &  \color{blue}{\textbf{50.9}} & 76.6  &	\color{blue}{\textbf{52.7}} & 17.5 & 35.7 & 24,069	& \color{red}{\textbf{250,768}} & \color{red}{\textbf{2,474}} & 5,317 &  18.3 \\	
FAMNet~\cite{ChuLin19} & \textbf{online}  & \color{red}{\textbf{52.0}} & 76.5 & 48.7 & \color{blue}{\textbf{19.1}} & \color{blue}{\textbf{33.4}} & \color{red}{\textbf{14,138}} & \color{blue}{\textbf{253,616}} & \color{blue}{\textbf{3,072}} & 5,318 &  0.6 \\
	
GMPHD-N1Tr~\cite{BaiWal19} & \textbf{online} & 42.1 & \color{red}{\textbf{77.7}} & 33.9 &	11.9 &	42.7 &	18,214	& 297,646 &	10,698	& 10,864 & 9.9\\
GMPHD-KCF~\cite{KutBosEis17} & \textbf{online} & 40.3 & 75.4 & 36.6 & 8.6 & 43.1 & 47,056 & 283,923 & 5,734 & 7,576 & 3.3\\
GMPHDOGM17~\cite{SonYooYow19} & \textbf{online} & 49.9 & 77.0 & 47.1 & \color{red}{\textbf{19.7}} & 38.0 & 24,024 & 255,277 & 3,125 & \color{red}{\textbf{3,540}} & \color{blue}{\textbf{30.7}} \\
PHD-DCM~\cite{FuSngNaq18} & \textbf{online} & 46.5 & 77.2 & N/A & 16.9 & 37.2 & 23,859 & 272,430 & 5,649 & 9,298 & 1.6 \\
GMPHD-SHA~\cite{SonJeo16} & \textbf{online} & 43.7	& 76.5 & 39.2 & 11.7 & 43.0 &	25,935 & 287,758 & 3,838 & \color{blue}{\textbf{5,056}} & 9.2\\
GMPHD-DAL~\cite{Nat19} & \textbf{online} & 44.4 & \color{blue}{\textbf{77.4}} & 36.2 & 14.9 & 39.4 & 19,170 & 283,380 & 11,137 & 13,900 & 3.4\\
\textbf{GMPHD-ReId (ours)} & \textbf{online} & 46.8 & 76.4 & \color{red}{\textbf{54.1}} & \color{red}{\textbf{19.7}} & \color{red}{\textbf{33.3}} & 38,452 & 257,678 & 3,865 & 8,097 & \color{red}{\textbf{30.8}} \\
\hline
\end{tabular}
\end{center}
\caption{Tracking performance of representative trackers developed using both online and offline methods. All trackers are evaluated on the test data set of the \textbf{MOT17} benchmark using public detections. The first and second highest values are highlighted by $\color{red}{\textbf{red}}$ and $\color{blue}{\textbf{blue}}$, respectively (for both online and offline trackers). Evaluation measures with ($\uparrow$) show that higher is better, and with ($\downarrow$) denote lower is better. N/A shows not available.}
\label{tbl:MOT17}
\end{table*}

\begin{figure*} [htb] 
  \begin{center}
  {\label{fig:DetectionFrame57} \includegraphics[height=0.272\textwidth]{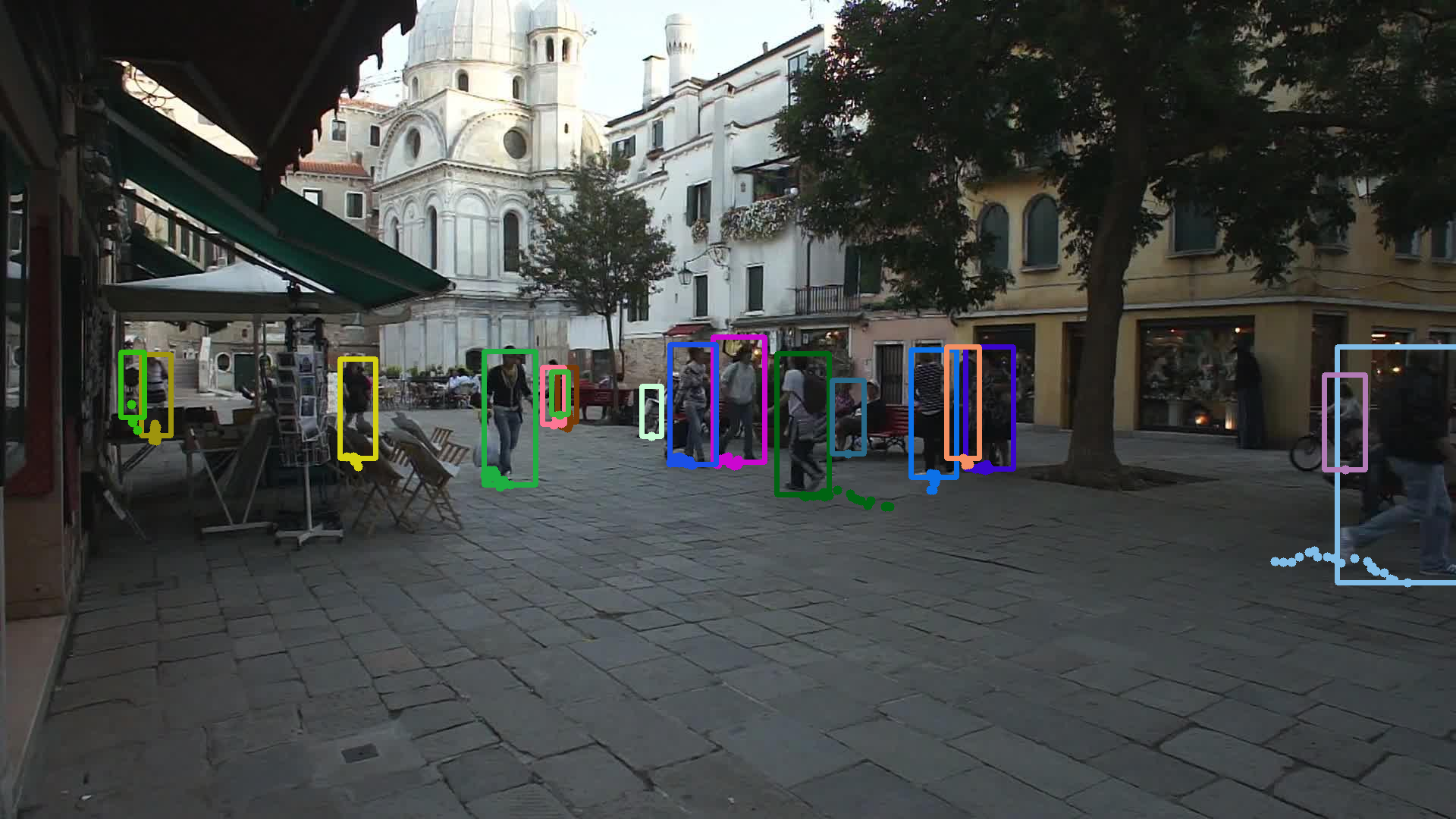}} 
  {\label{fig:TriPHDFrame57} \includegraphics[height=0.272\textwidth]{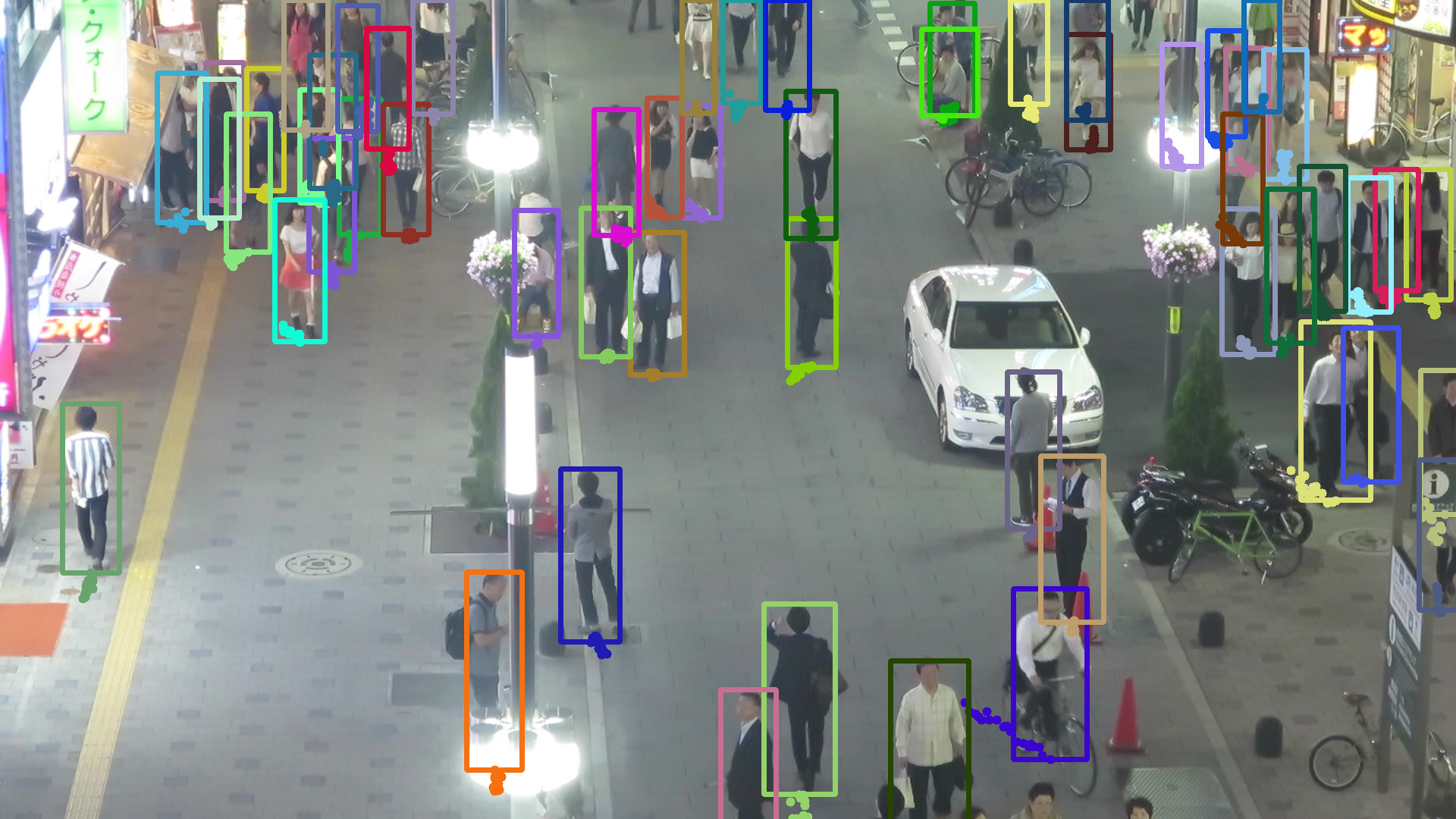}} \\
  {\label{fig:ThreePHDsFrame57} \includegraphics[height=0.31\textwidth]{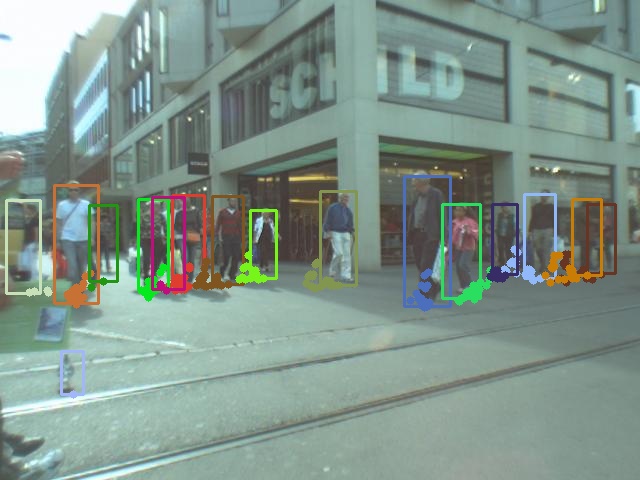}}  
  {\label{fig:ThreePHDsFrame57} \includegraphics[height=0.31\textwidth]{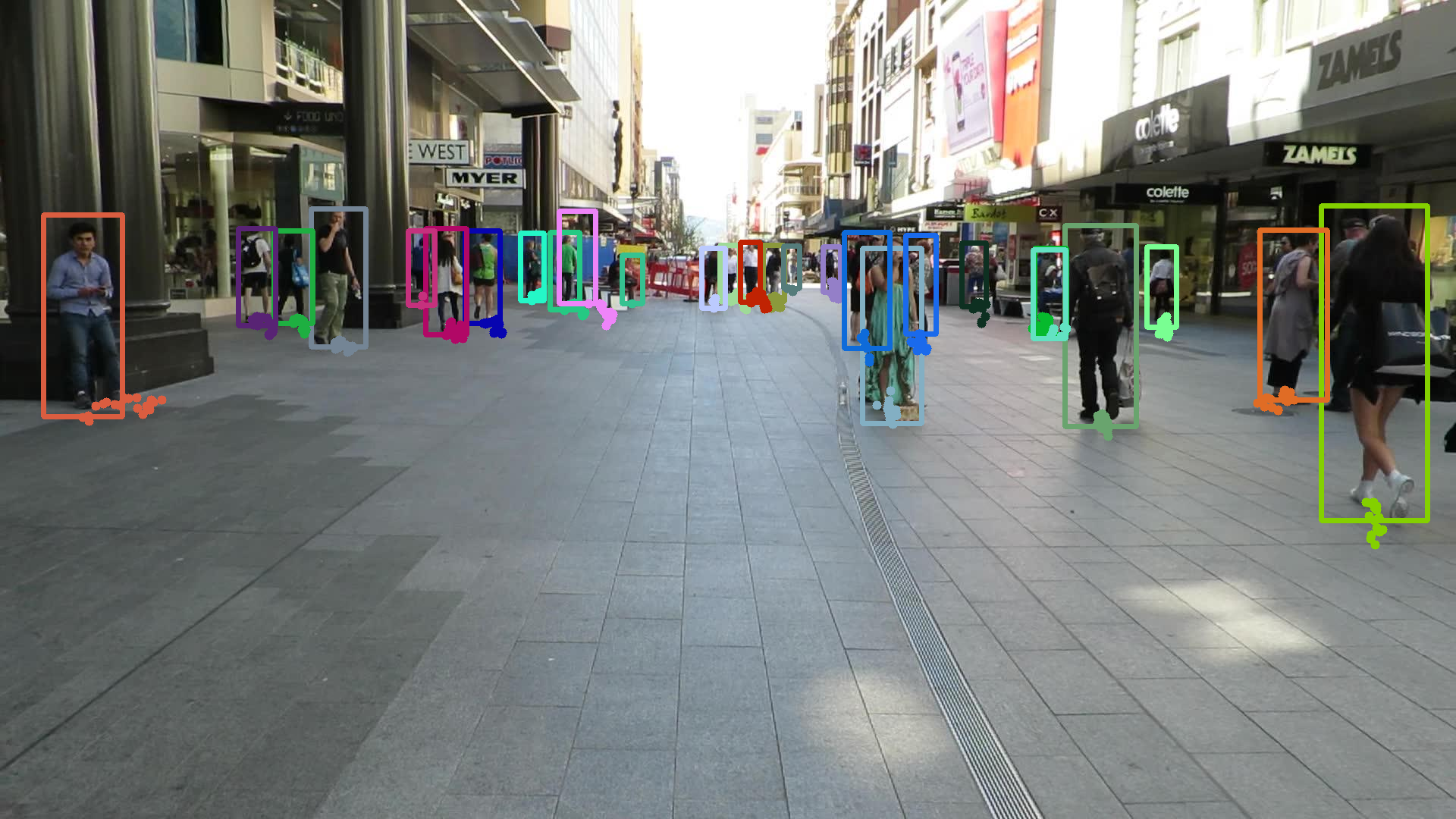}}\\ 
  {\label{fig:ThreePHDsFrame57} \includegraphics[height=0.273\textwidth]{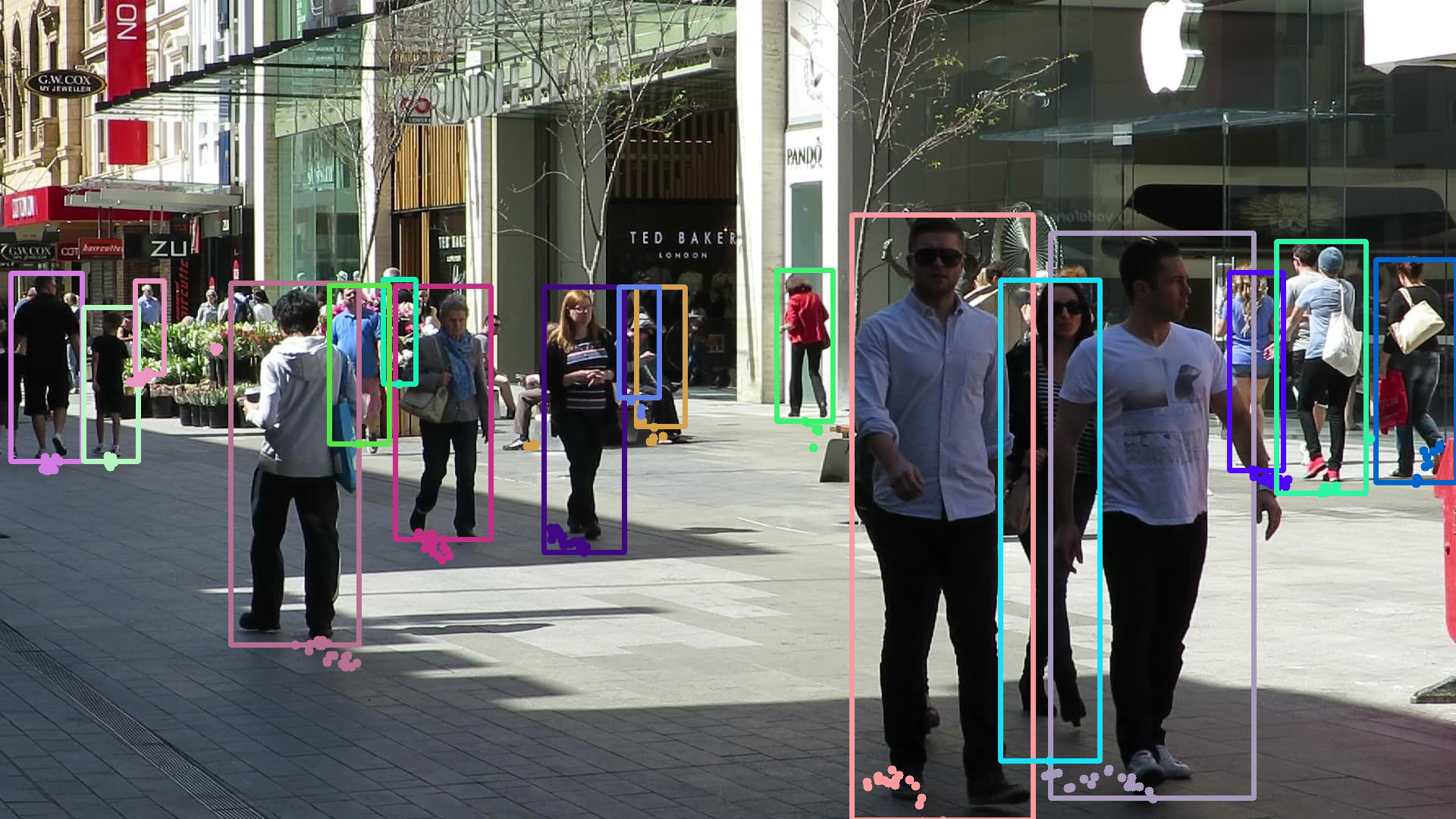}} 
  {\label{fig:ThreePHDsFrame57} \includegraphics[height=0.273\textwidth]{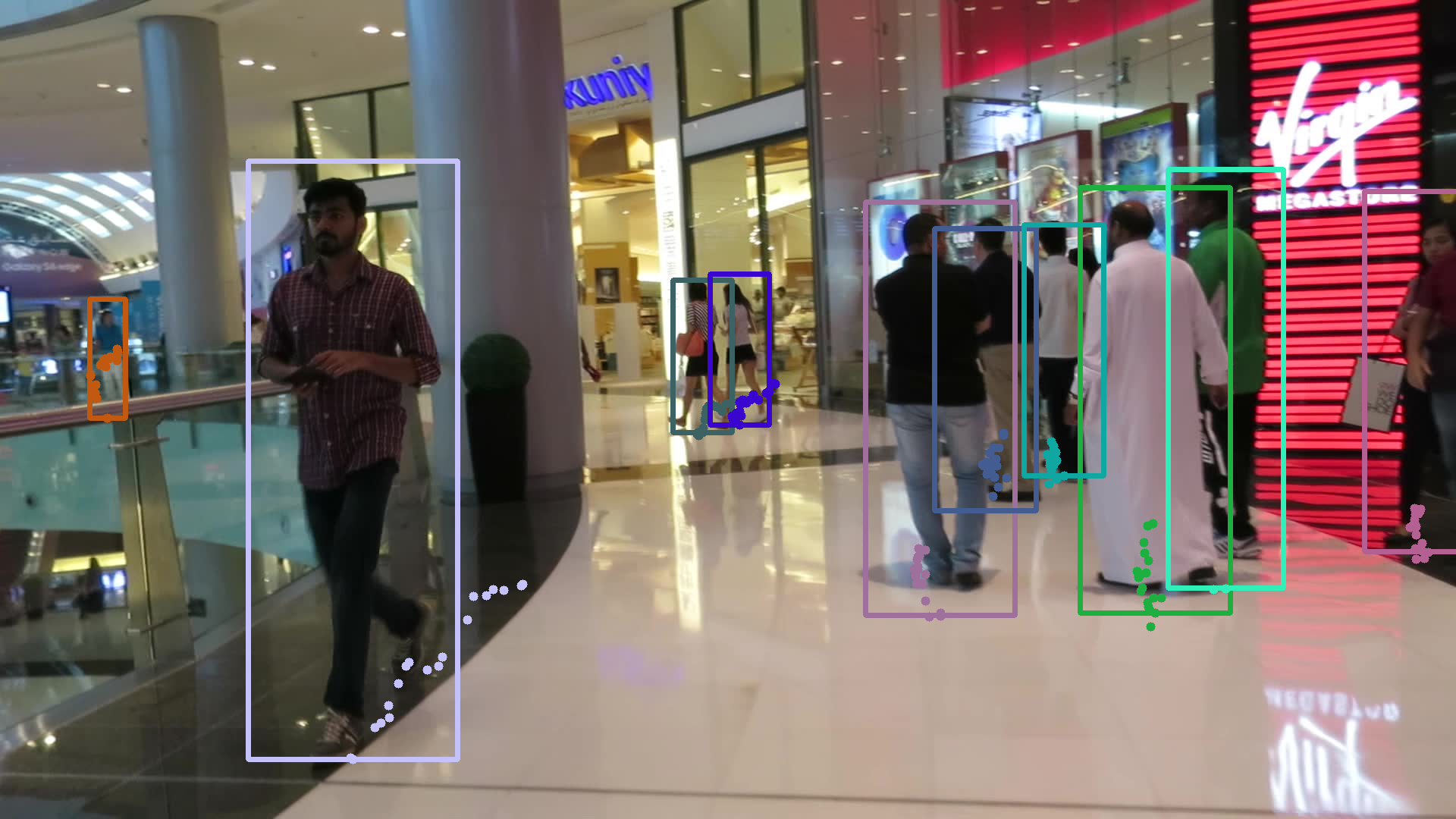}} \\
  {\label{fig:ThreePHDsFrame57} \includegraphics[height=0.38\textwidth]{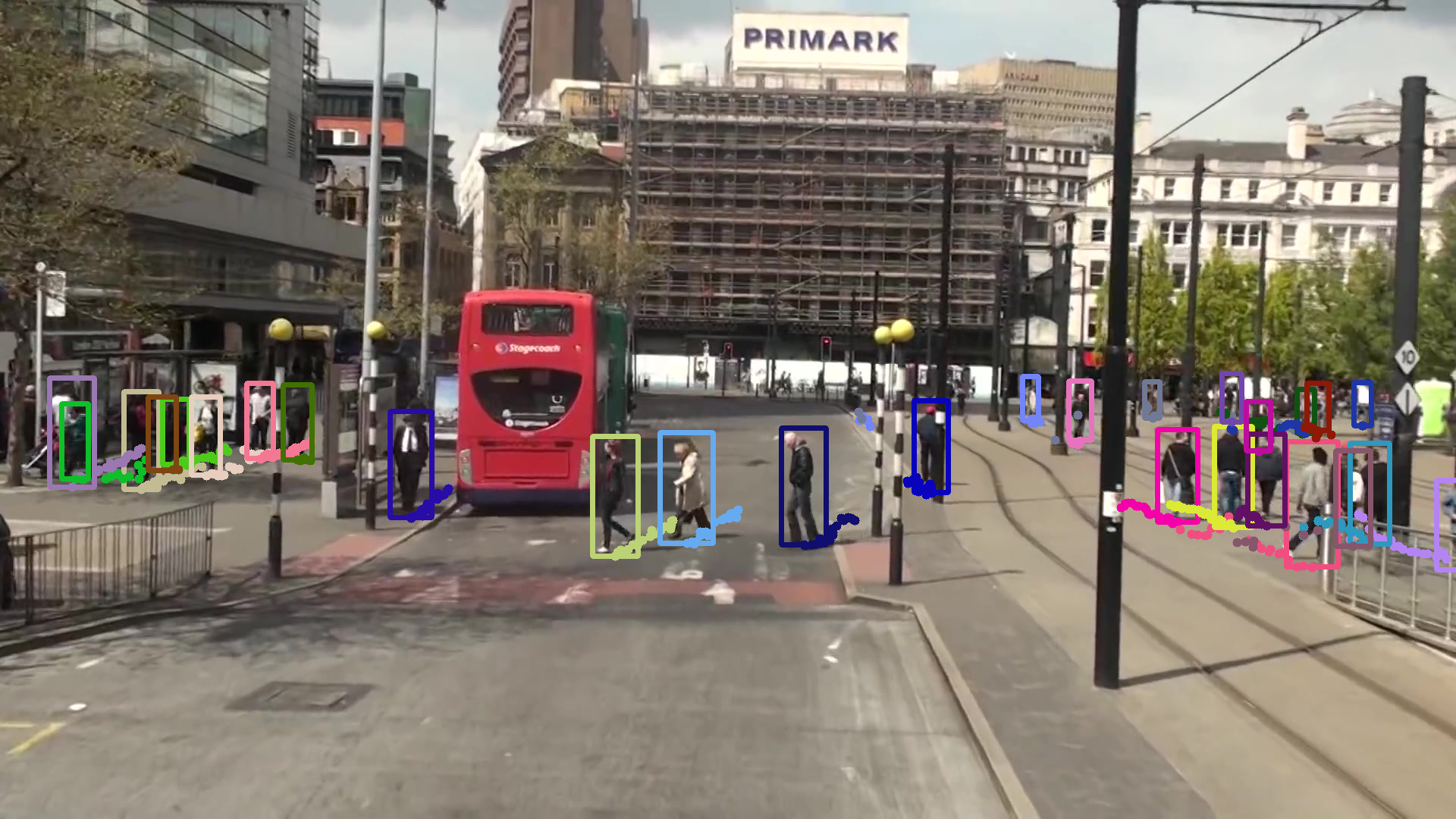}}  
  \end{center}
   \caption{Sample results on several test sequences of MOT17 data sets using SDP detector; bounding boxes
represent the tracking results with their color-coded identities. From left to right: MOT17-01-SDP and MOT17-03-SDP (top row), MOT17-06-SDP and MOT17-07-SDP (the 1st middle row), MOT17-08-SDP and MOT17-12-SDP (the 2nd middle row), and MOT17-14-SDP (bottom row). The videos of tracking results are available on the MOT Challenge website \color{red}{https://motchallenge.net/}.}
  \label{fig:MOTA17}
\end{figure*}

Sample qualitative tracking results are shown as examples in Fig.~\ref{fig:MOTA17} using SDP detector and MOT17 test sequences. The tracking results are represented by bounding boxes and short trajectories with their color-coded identities. On the top row, MOT17-01-SDP and MOT17-03-SDP are shown from left to right. In the first and second middle rows are MOT17-06-SDP and MOT17-07-SDP, and MOT17-08-SDP and MOT17-12-SDP, respectively. Finally, MOT17-14-SDP is shown on the bottom row.

\noindent

\subsection{Evaluations on HiEve Benchmark Test Sets}

We also evaluate our proposed tracker on the HiEve~\cite{LinLiuLiu21} benchmark using the provided public detections with similar parameters setting as our evaluations on the MOT benchmark given above. There are two additional evaluation metrics, w-MOTA and IDSw-DT, introduced for HiEve dataset. The w-MOTA is computed in a similar manner as MOTA except that a higher weight is assigned to the ID switch cases happening in disconnected tracks. The IDSw-DT is the total number of identity switches (IDSw) happening in disconnected tracks whose value will be used when computing the w-MOTA. The quantitative evaluations of our proposed tracker is compared with other state-of-the-art trackers in Table~\ref{tbl:HiEve}. As shown in this table, our tracker outperforms all online trackers in MOTA, w-MOTA, MT, ML, FN and Frag evaluation metrics. It is also ranked 2nd in IDF1 and IDSw-DT. Interestingly, our tracker outperforms all other trackers, both online and offline, not only those listed in the Table~\ref{tbl:HiEve} but also all those listed in the HiEve leaderboard in MT, ML and FN evaluation metrics, while running in real-time.

\begin{table*}[htbp]
\begin{center}
\begin{tabular}{|l|c|c|c|c|c|c|c|c|c|c|c|r|}
\hline
Tracker & Mode & MOTA$\uparrow$ & w-MOTA$\uparrow$ & MOTP$\uparrow$ & IDF1$\uparrow$ & MT (\%)$\uparrow$ & ML (\%)$\downarrow$ & FP$\downarrow$ & FN$\downarrow$ & IDSw$\downarrow$ & IDSw-DT $\downarrow$ & Frag$\downarrow$ \\
\hline
LinkBox~\cite{JinYueYab20}  & \textbf{offline} & \color{red}{\textbf{51.38}} & \color{red}{\textbf{46.56}} & 76.41 & 47.15 & \color{red}{\textbf{29.28}} &	\color{red}{\textbf{29.07}} &	\color{red}{\textbf{2804}} & \color{red}{\textbf{29345}} &	1725 & 	\color{red}{\textbf{84}} & \color{red}{\textbf{1645}} 	\\
SelectiveJDE~\cite{WuLinChe20}  & \textbf{offline} & \color{blue}{\textbf{50.59}} & \color{blue}{\textbf{45.42}} & \color{blue}{\textbf{76.79}} &	\color{red}{\textbf{56.81}} &	25.08 &	30.33 &	\color{blue}{\textbf{2860}} & \color{blue}{\textbf{29850}} & \color{blue}{\textbf{1719}} &	\color{blue}{\textbf{90}} & 2993 	\\
FCS-Track~\cite{ShuBerWan20}  & \textbf{offline} & 47.81 &	42.52 &	\color{red}{\textbf{76.85}} & \color{blue}{\textbf{49.82}} & \color{blue}{\textbf{25.29}} &	30.22 &	3847 & 30862 & 	\color{red}{\textbf{1658}} & 	92 & \color{blue}{\textbf{2625}}  \\
STPP~\cite{TaoKeaWei20}  & \textbf{offline} &  37.47 &	32.07 &	75.60 & 40.17 &	20.36 & \color{blue}{\textbf{29.80}} & 7395 & 31638 & 4536 & 94 &	3486 \\	
TPM~\cite{JinTaoWei20}  & \textbf{offline} &  33.58 &	28.30 &	75.58 & 37.71 &	10.70 &	31.17 & 6595 & 	35395 &	4287 & 	92 & 6070 	\\  
\hline
DeepSORT~\cite{WojBewPau17} & \textbf{online} & 27.12 &	21.95 & 70.47 & 28.55 & 8.50 & 	41.45 & \color{red}{\textbf{5894}} & 	42668  & \color{blue}{\textbf{2220}} & 	90 & 	\color{blue}{\textbf{3122}} \\ 
CenterTrack~\cite{ZhoKolKra20} & \textbf{online}  & \color{blue}{\textbf{31.06}} & \color{blue}{\textbf{25.66}} & \color{blue}{\textbf{75.77}} & \color{red}{\textbf{41.81}} &	8.60 & \color{blue}{\textbf{27.91}} & 10014 &	\color{blue}{\textbf{35253}} &	2767 &	94 &	7908  \\
MOTDT~\cite{CheAiZhu18} & \textbf{online} & 26.09 & 21.73 & \color{red}{\textbf{76.50}} & 32.88 & \color{blue}{\textbf{8.70}} & 54.56 & \color{blue}{\textbf{6318}} & 43577 & \color{red}{\textbf{1599}} & \color{red}{\textbf{76}}  & N/A \\	

\textbf{GMPHD-ReId} &  &   &  &  &  &  &  &  &  &	 &  &	  \\
\textbf{(ours)} & \textbf{online} &  \color{red}{\textbf{31.31}} & \color{red}{\textbf{26.20}} & 75.27 & \color{blue}{\textbf{37.68}} & \color{red}{\textbf{35.99}} & \color{red}{\textbf{24.34}} & 17309 & \color{red}{\textbf{26158}} &	4392 & \color{blue}{\textbf{89}} &	\color{red}{\textbf{3063}}  \\
\hline
\end{tabular}
\end{center}
\caption{Tracking performance of representative trackers developed using both online and offline methods. All trackers are evaluated on the test data set of the \textbf{HiEve}~\cite{LinLiuLiu21} benchmark using public detections. The first and second highest values are highlighted by $\color{red}{\textbf{red}}$ and $\color{blue}{\textbf{blue}}$, respectively (for both online and offline trackers). Evaluation measures with ($\uparrow$) show that higher is better, and with ($\downarrow$) denote lower is better. N/A shows not available.}
\label{tbl:HiEve}
\end{table*}

\section{Conclusions} \label{Sec:Conclusions}

We have developed a novel online multi-target visual tracker based on the GM-PHD filter and deep CNN appearance representations learning which runs in real-time; it is suitable for real-time applications such as autonomous driving and can be extended for tracking targets in a camera network due to its computational efficiency. We apply this method for tracking multiple targets in video sequences acquired under varying environmental conditions and targets density. We followed a tracking-by-detection approach using the public detections provided in the MOT16, MOT17 and HiEve benchmark data sets. We integrate spatio-temporal similarity from the object bounding boxes and the appearance information from the learned deep CNN (using both motion and appearance cues) to label each target in consecutive frames. We also formulate an augmented likelihood using these two information sources and then integrate into the update step of the GM-PHD filter. We learn the deep CNN appearance representations by training an identification network (IdNet) on large-scale person re-identification data sets. We also employ additional unassigned tracks prediction after the GM-PHD filter update step to overcome the susceptibility of the GM-PHD filter towards miss-detections caused by occlusion. Results show that our method outperforms state-of-the-art trackers developed using both online and offline approaches on the MOT16, MOT17 and HiEve benchmark data sets in terms of tracking accuracy and identification. In the future work, we will include inter-object relations model for tackling the interactions of different objects.

\bibliographystyle{IEEEtran}
\bibliography{egbib}

\end{document}